\documentclass[letterpaper, 10 pt, conference]{ieeeconf}  

\IEEEoverridecommandlockouts                              

\usepackage{amsmath,amssymb,amsfonts}
\usepackage{graphicx} 
\usepackage[dvipsnames]{xcolor}
\usepackage{algorithm}
\usepackage{booktabs} 
\usepackage{algpseudocode}
\usepackage{multirow}
\usepackage{svg}
\usepackage{subcaption}
\usepackage{xspace}
\usepackage{xcolor}
\usepackage{colortbl}
\usepackage{tikz}
\usepackage{pifont}
\usepackage{gensymb}

\colorlet{tabfirst}{Green!35}
\definecolor{tabthird}{rgb}{1, 0.85, 0.7}
\definecolor{tabsecond}{rgb}{1, 0.96, 0.7}

\newcommand{\first}[1]{\textcolor{ForestGreen}{\textbf{#1}}}

\newcommand{\greencheck}{{\color{ForestGreen}\checkmark}}
\newcommand{\redx}{{\color{red}\ding{55}}}

\newcommand{\ours}{\text{TransLocNet}\xspace}

\makeatletter
\let\NAT@parse\undefined
\makeatother
\usepackage{hyperref}
\hypersetup{
    colorlinks=true,
    linkcolor=blue,
    filecolor=magenta,      
    urlcolor=cyan,
    citecolor=red,
    pdftitle={goslam},
    pdfpagemode=FullScreen,
}

\title{\LARGE \bf
\ours: Cross-Modal Attention for Aerial-Ground Vehicle Localization with Contrastive Learning
}

\author{Phu Pham$^{1}$, Damon Conover$^{2}$,  Aniket Bera$^{1}$\\
$^1$Department of Computer Science, Purdue University  $^2$DEVCOM Army Research Laboratory\\
\texttt{\{phupham, aniketbera\}@purdue.edu, damon.m.conover.civ@army.mil}
}

\begin{document}

\maketitle
\thispagestyle{empty}
\pagestyle{empty}




\begin{abstract}

Aerial–ground localization is difficult due to large viewpoint and modality gaps between ground-level LiDAR and overhead imagery. We propose \ours, a cross-modal attention framework that fuses LiDAR geometry with aerial semantic context. LiDAR scans are projected into a bird’s-eye-view representation and aligned with aerial features through bidirectional attention, followed by a likelihood map decoder that outputs spatial probability distributions over position and orientation. A contrastive learning module enforces a shared embedding space to improve cross-modal alignment. Experiments on CARLA and KITTI show that \ours outperforms state-of-the-art baselines, reducing localization error by up to 63\% and achieving sub-meter, sub-degree accuracy. These results demonstrate that \ours provides robust and generalizable aerial–ground localization in both synthetic and real-world settings.

\end{abstract}

\section{INTRODUCTION}


Vehicle localization is essential for autonomous navigation, advanced driver assistance systems, and mobile robotics \cite{Levinson-RSS-07, survey_laconte}. Conventional approaches often rely on Global Navigation Satellite System (GNSS)-based methods such as GPS/INS. However, these solutions degrade in GNSS-denied environments like urban areas, tunnels, and dense forests, where occlusion, multipath, and jamming can introduce multi-meter errors \cite{Jarraya2025}. To address these issues, researchers have explored sensor-based alternatives such as visual odometry and Simultaneous Localization and Mapping (SLAM). These methods provide relative pose estimates but accumulate drift over long distances without global references \cite{SLAM-Lategahn, SLAM-review-cheng}. High-definition (HD) maps offer absolute localization by matching sensor data to compact semantic representations \cite{HDmap-Soorchaei, hdmap-gong}, but their construction and maintenance at city scale are costly, requiring frequent re-collection and annotation.

\begin{figure}
    \centering
    \includegraphics[width=\linewidth]{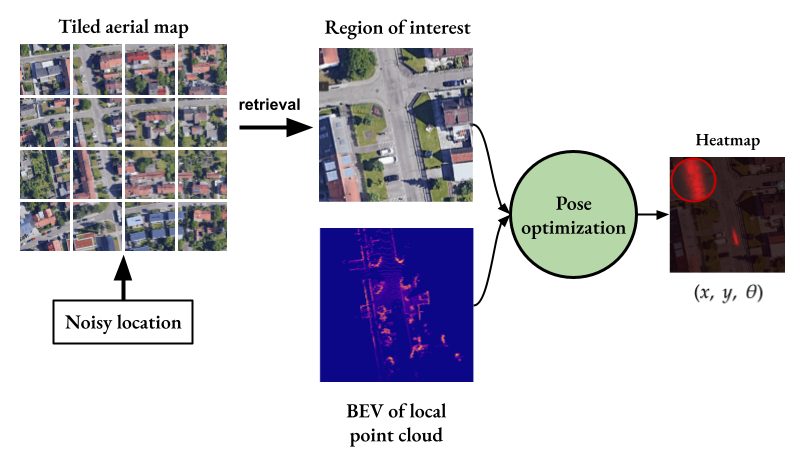}
    \caption{\small Overview of our aerial-ground vehicle localization framework. Starting from a noisy initial position, a candidate region is retrieved from a tiled aerial map. The local LiDAR BEV projection is then jointly optimized with the aerial image patch to refine position $(x, y)$ and orientation $\theta$.}
    \label{fig:teaser}
\end{figure}

Overhead imagery has emerged as an alternative global map source \cite{HighlyAccurate-shi, wang2022satellite}. Cross-view matching aligns ground-level sensor observations with aerial views to infer ego-vehicle pose. CVM-Net \cite{CVM-Net} introduced cross-view matching networks for viewpoint-invariant descriptors, while HighlyAccurate \cite{HighlyAccurate-shi} formulated localization as direct pose estimation using CNNs and geometric projection. Transformer-based approaches \cite{layer2layer-yang} improved robustness by modeling spatial relationships across views.

Most existing methods focus on visual data and underutilize LiDAR, which provides dense 3D structure resilient to illumination and weather variations \cite{yang2025evaluating}. LiDAR complements aerial imagery with ground-level geometry, but prior fusion strategies typically rely on feature concatenation, which fails to capture spatial correspondences. They also lack robustness to large viewpoint shifts, scale discrepancies, and temporal misalignment. Attention mechanisms and contrastive learning have been explored independently for cross-modal alignment \cite{contrastive-coding, contrastive-thoma, deng2025spatiotemporal}, but have not been jointly applied to aerial–ground localization.

\begin{figure*}
    \centering
    \includegraphics[width=\linewidth]{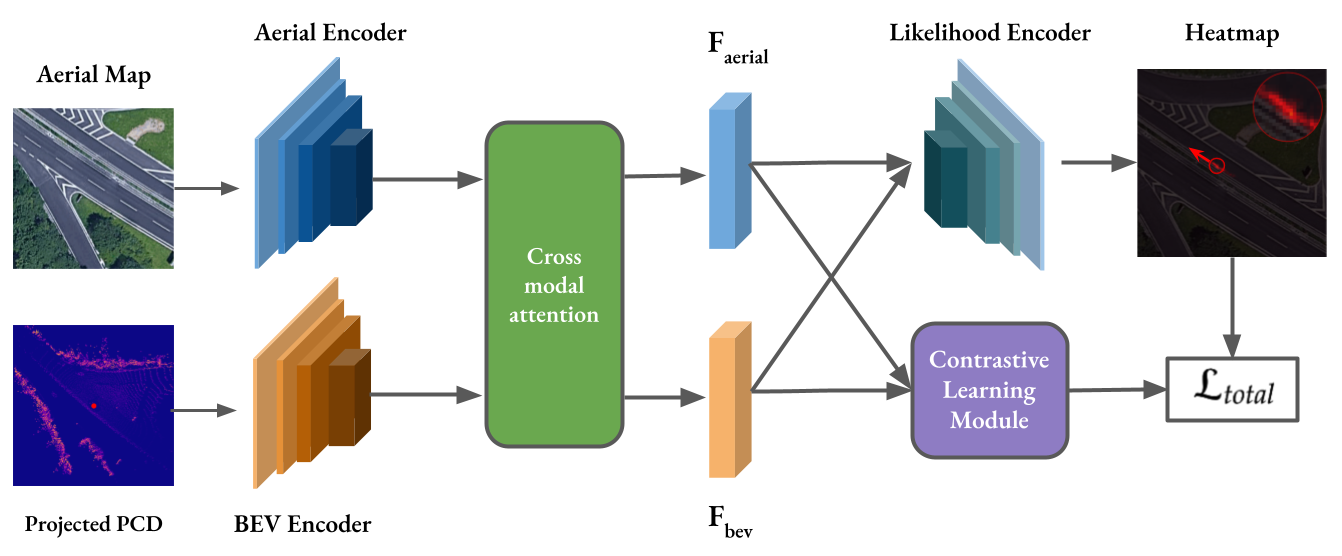}
    \caption{\small \textbf{Overview:} \ours localizes aerial and ground vehicles by encoding aerial images and LiDAR BEV projections into respective feature maps. A cross-modal attention aligns features, which are fused and decoded into likelihood maps for uncertain position and orientation estimates. Contrastive learning improves alignment by bringing matched aerial-BEV pairs closer and pushing mismatched pairs apart.}
    \label{fig:architecture}
\end{figure*}

Despite these advancements, existing approaches face limitations in handling large-scale viewpoint shifts, modality discrepancies, and noisy real-world data, often relying on simplistic fusion strategies that fail to capture intricate cross-modal interactions. To address these gaps, we propose \ours, a novel framework for aerial-ground vehicle localization that integrates cross-modal attention mechanisms with contrastive learning. By projecting local LiDAR point clouds to a top-view representation and aligning them with aerial images, \ours employs attention-based fusion to model inter-modal dependencies and a contrastive loss to enforce semantic alignment, achieving robust pose estimation. An overview of our proposed framework is illustrated in Fig.~\ref{fig:teaser}.

The main contributions of this work are threefold: 
\begin{itemize}
    \item We propose a cross-modal attention module that dynamically attends to salient features across bird's-eye-view (BEV) point clouds and aerial imagery, enabling robust feature fusion and improved localization accuracy under large viewpoint variations.  
    \item We integrate a contrastive learning objective that optimizes feature embeddings across modalities, strengthening retrieval and refining metric pose estimation.  
    \item We provide extensive empirical validation on both synthetic (CARLA \cite{guan2024agl}) and real-world (KITTI \cite{KITTI}) benchmarks, where our model achieves better performance, reducing location error by up to \textbf{63\%} and orientation error by up to \textbf{81\%} compared to prior methods.  

\end{itemize}

\section{RELATED WORK}

\subsection{Map-based localization without GNSS}

Classical autonomous driving systems localize against pre-built maps such as high-definition (HD) LiDAR reflectivity, semantic lane maps, or OpenStreetMap (OSM), typically using particle-filter or scan-matching pipelines. Landmark examples include HD-map particle filtering for urban driving, as well as a large body of Monte-Carlo/box-particle formulations that incorporate map-based priors \cite{levinson2007mapbased}. When HD maps are not available, lightweight OSM data can be fused with onboard sensors to help a robot recover its global position \cite{zhou2021osm, sarlin2023orienternet}. This is often achieved with learned image-to-map embeddings or constrained particle filtering.

\subsection{SLAM-based localization}

Simultaneous Localization and Mapping (SLAM) supports real-time map construction and pose estimation through onboard sensors. Classical visual SLAM pipelines \cite{mur2015orb,lsd-slam,teed2021droid} are well-established, while recent 3D Gaussian Splatting (3DGS) \cite{3dgs} methods \cite{monogs,yugay2023gaussian,loopsplat, pham2024flashslam} enable real-time mapping and high-quality rendering, though primarily for indoor environments.

SLAM systems suffer from drift accumulation and require consecutive frames for tracking, making global localization difficult without GPS or global priors. This motivates cross-modal localization, where aerial imagery serves as a valuable complement to traditional SLAM.

\begin{figure*}[htbp]
\centering
\tikzset{every picture/.style={line width=0.75pt}} 

\scalebox{0.8}{
\begin{tikzpicture}[x=0.75pt,y=0.75pt,yscale=-1,xscale=1]

\draw (85,105) node  {\includegraphics[width=97.5pt,height=97.5pt]{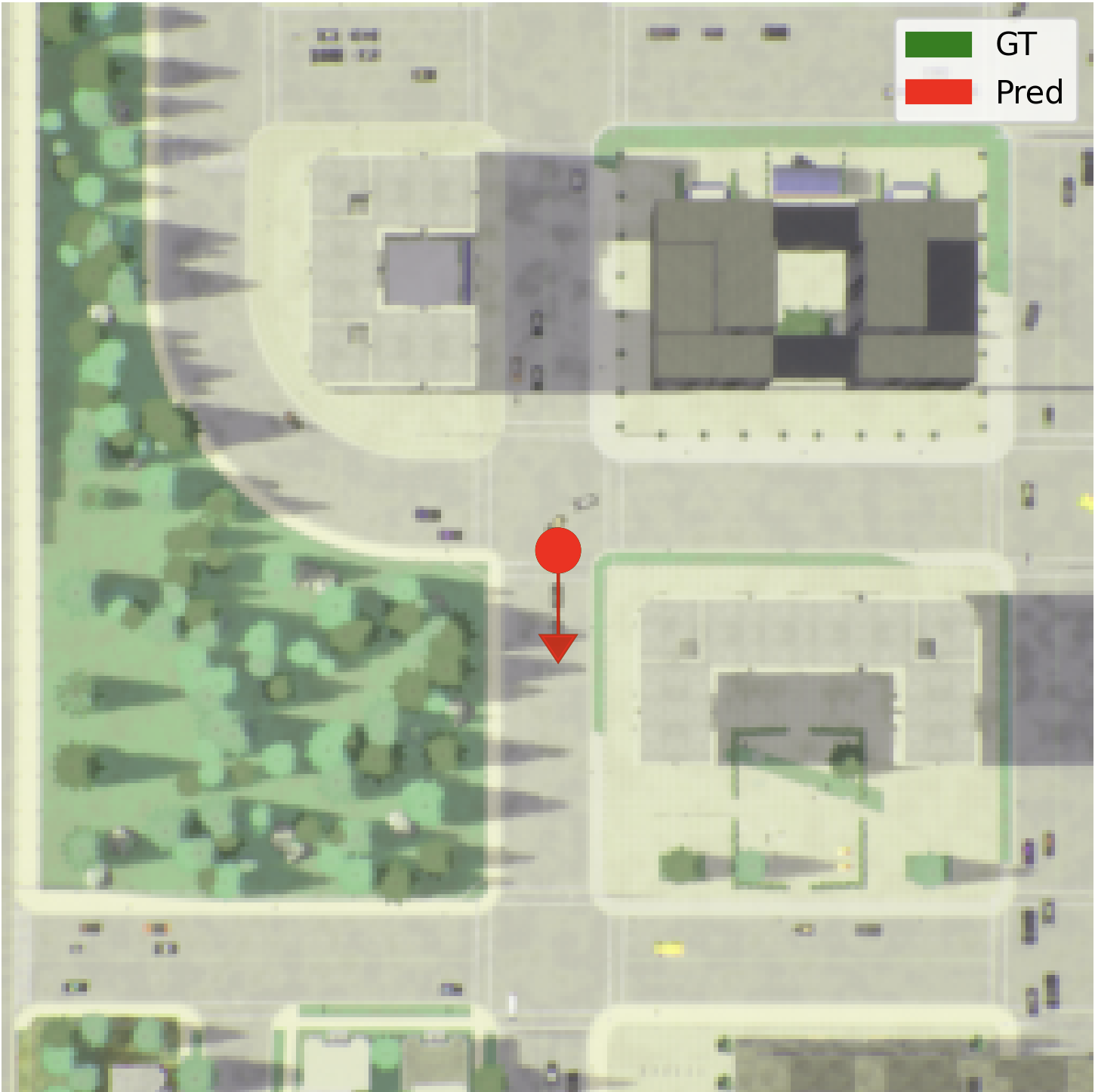}};
\draw (225,105) node  {\includegraphics[width=97.5pt,height=97.5pt]{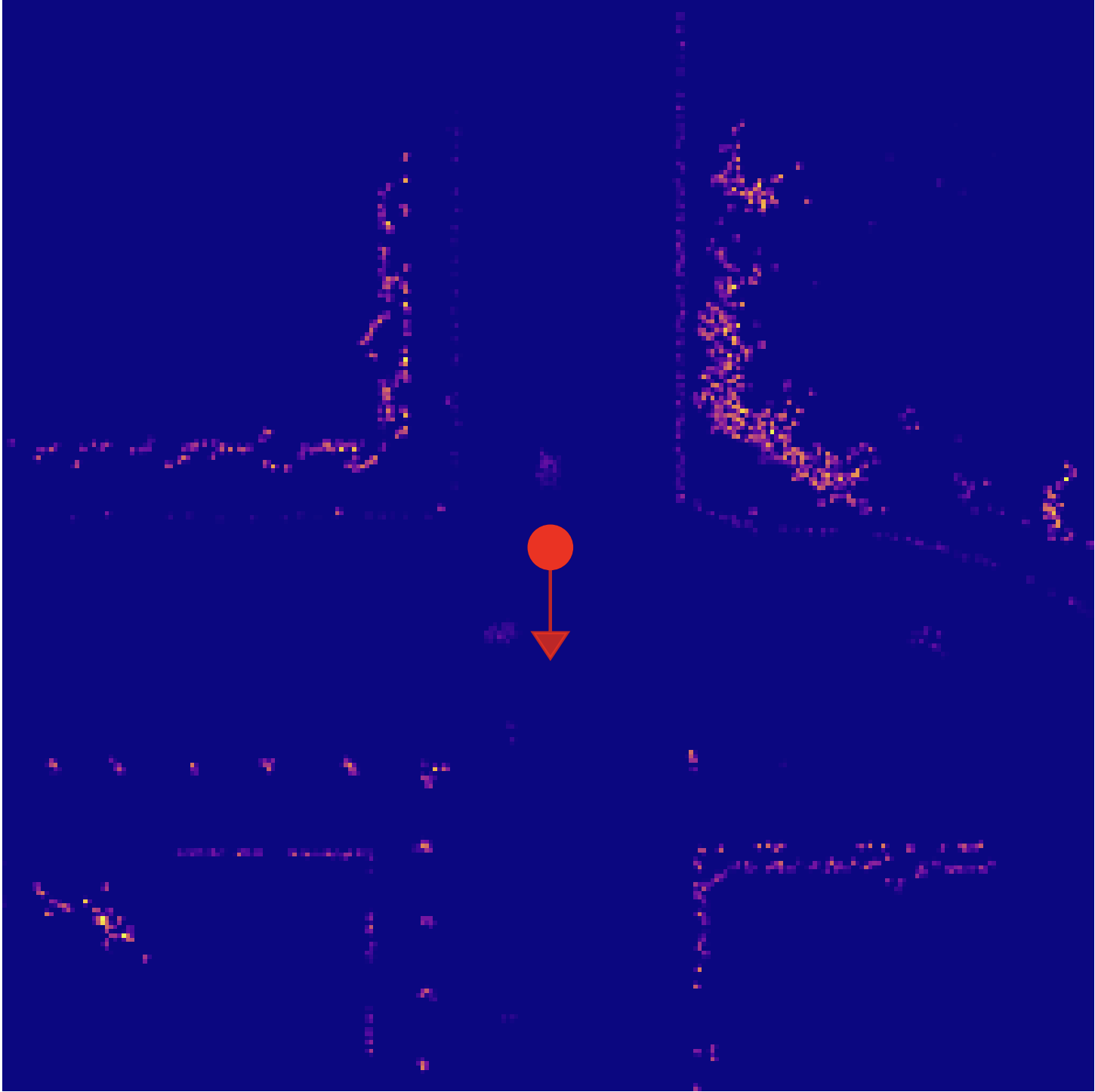}};
\draw (365,105) node  {\includegraphics[width=97.5pt,height=97.5pt]{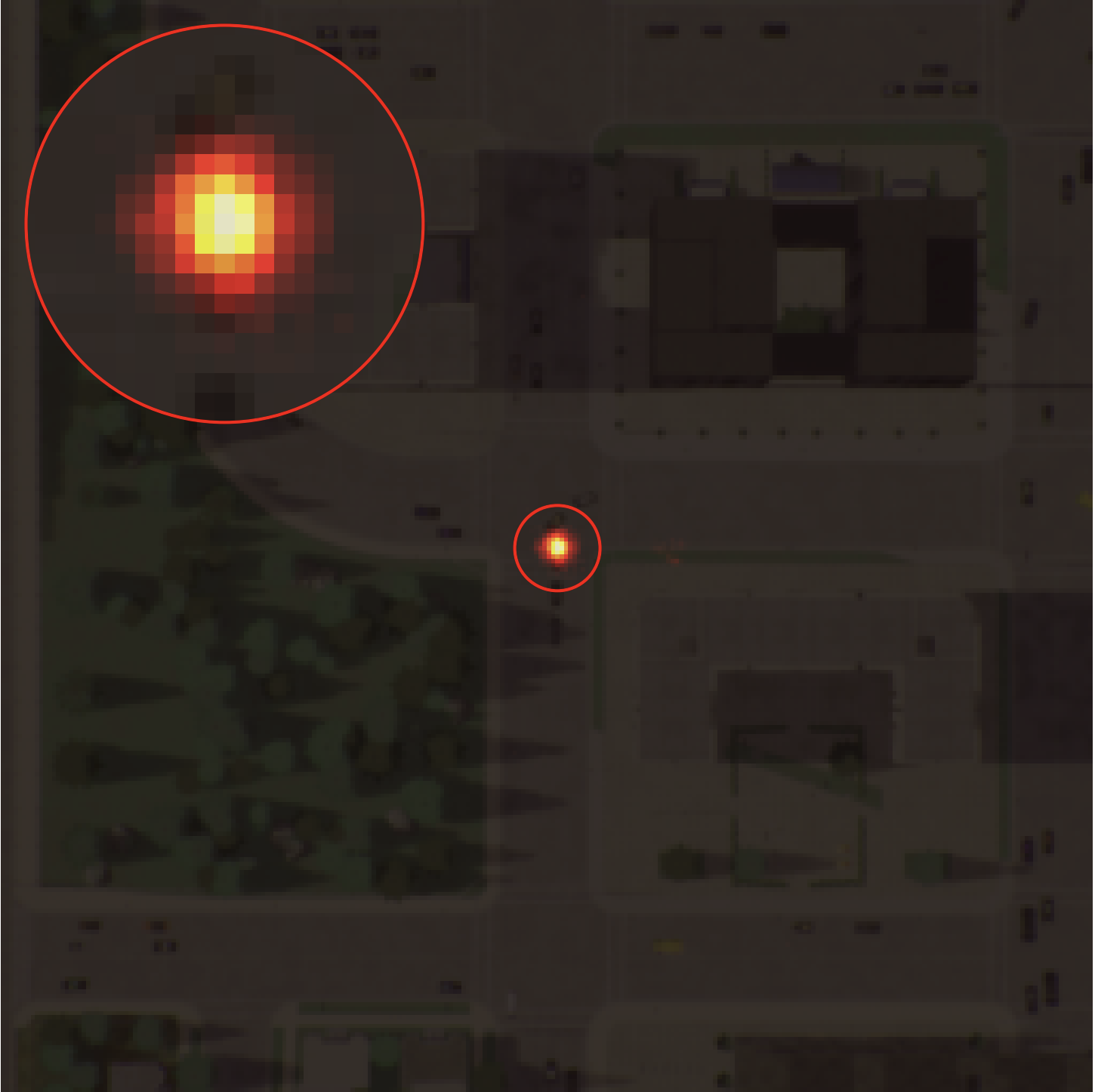}};
\draw (85,265) node  {\includegraphics[width=97.5pt,height=97.5pt]{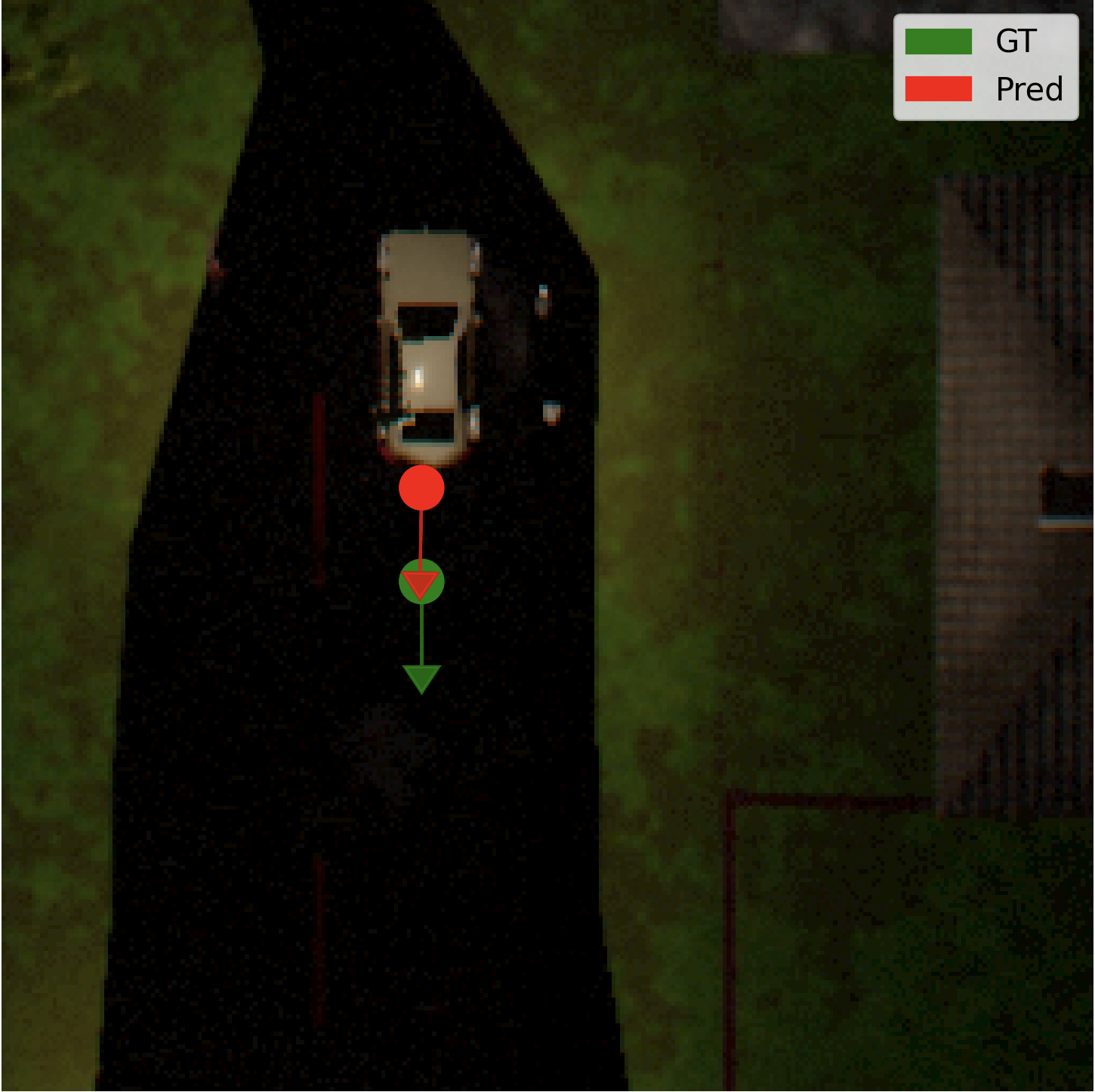}};
\draw (225,265) node  {\includegraphics[width=97.5pt,height=97.5pt]{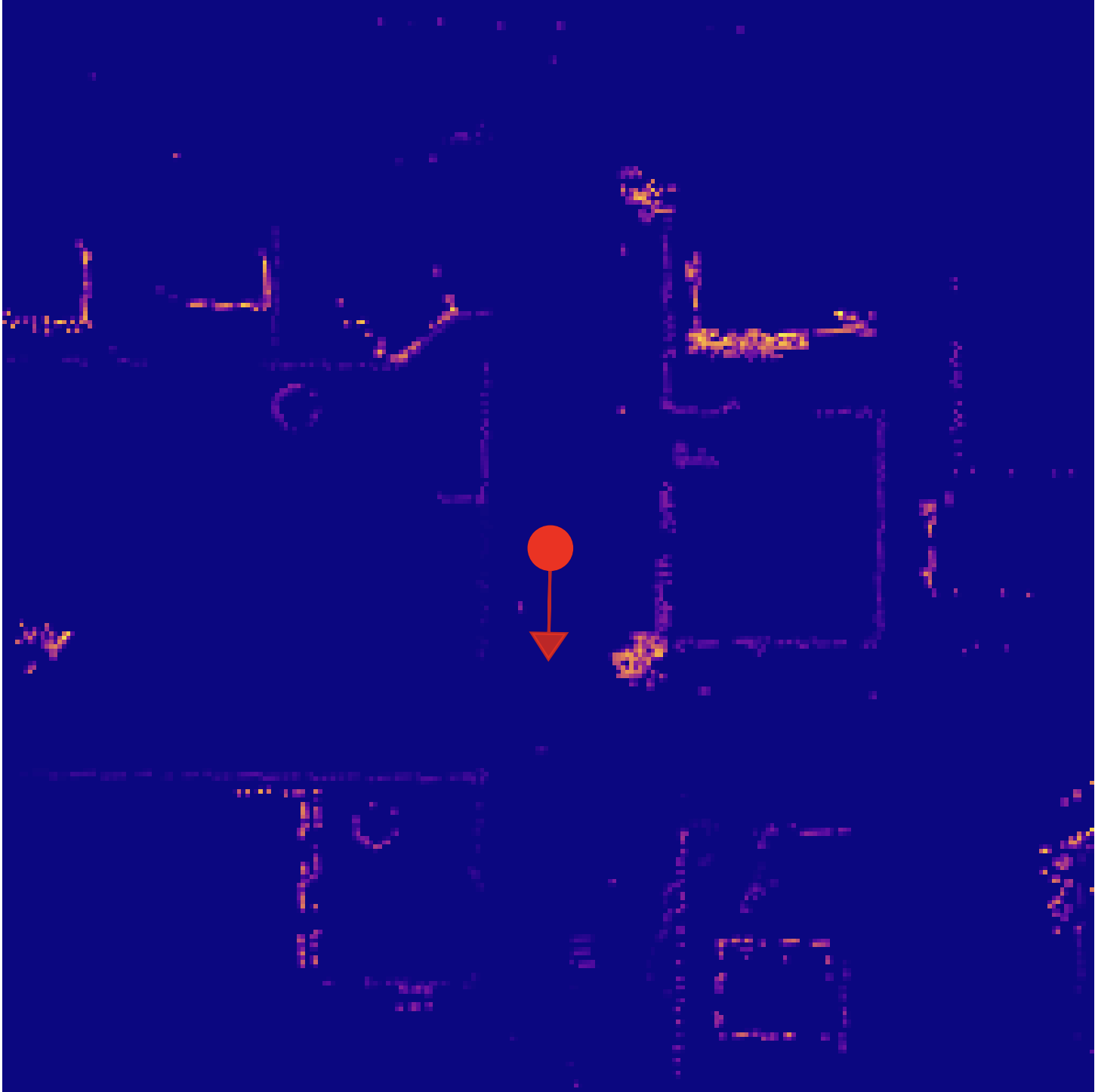}};
\draw (365,265) node  {\includegraphics[width=97.5pt,height=97.5pt]{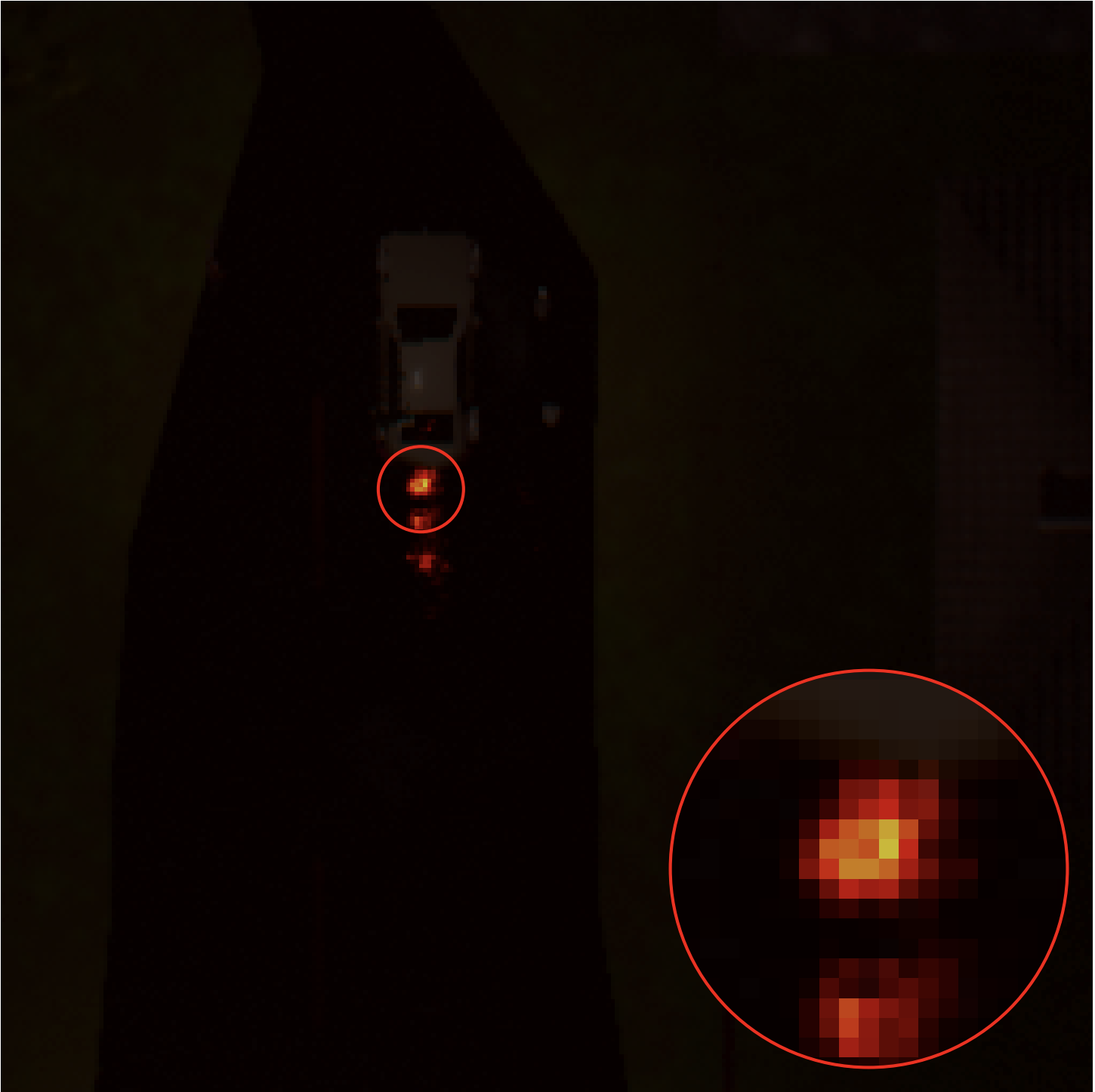}};
\draw (505,104) node  {\includegraphics[width=97.5pt,height=97.5pt]{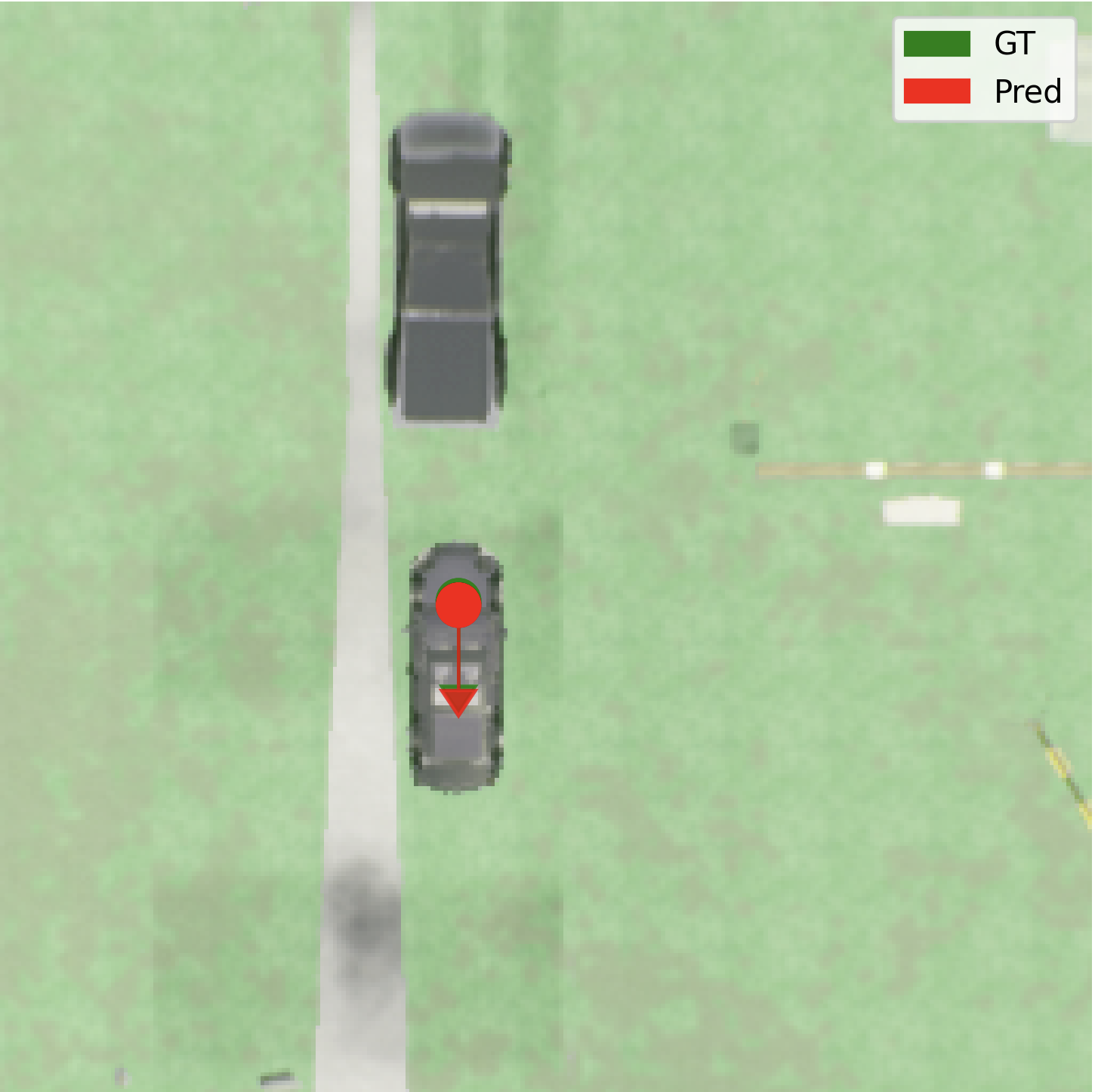}};
\draw (645,104) node  {\includegraphics[width=97.5pt,height=97.5pt]{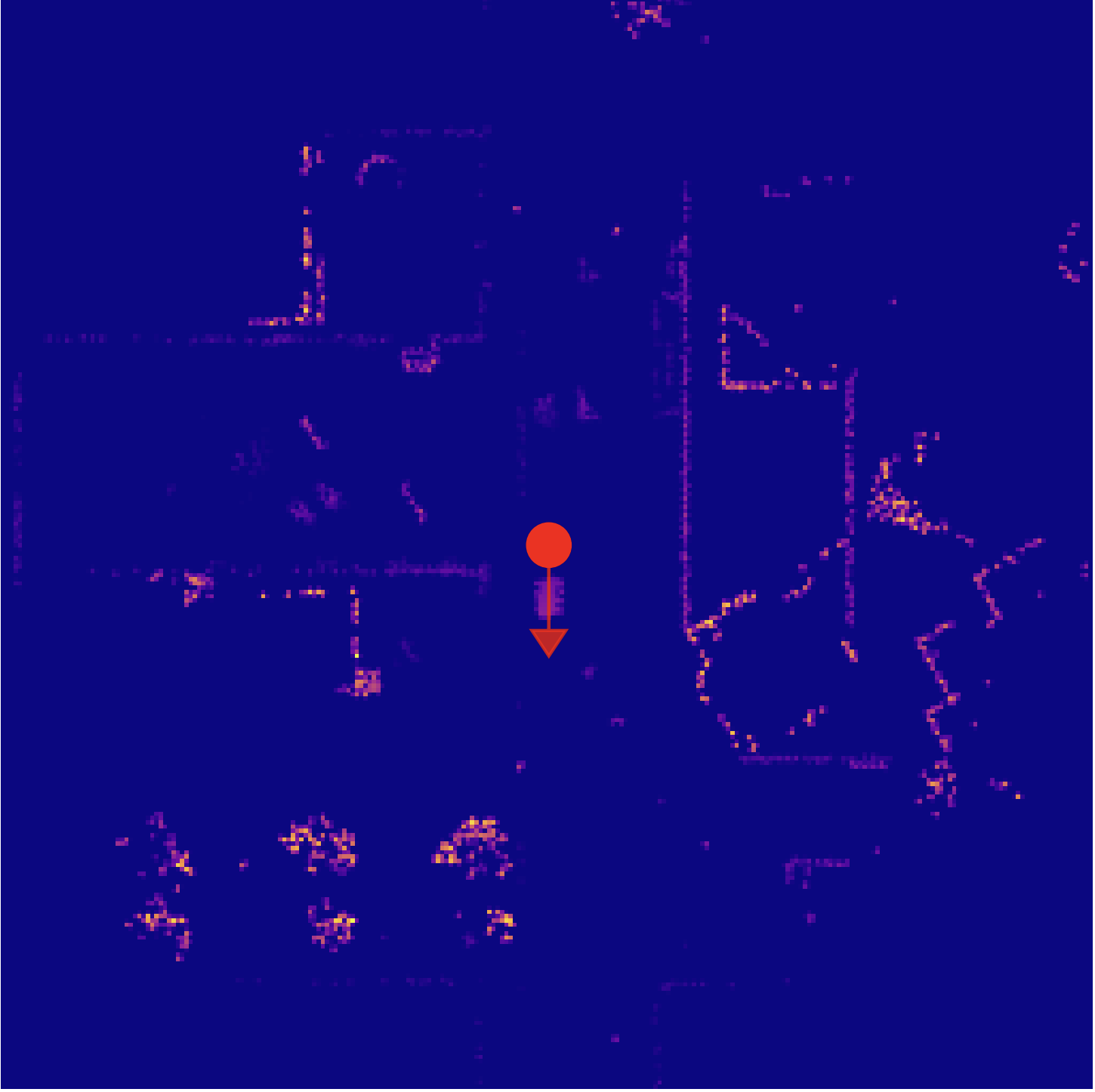}};
\draw (785,104) node  {\includegraphics[width=97.5pt,height=97.5pt]{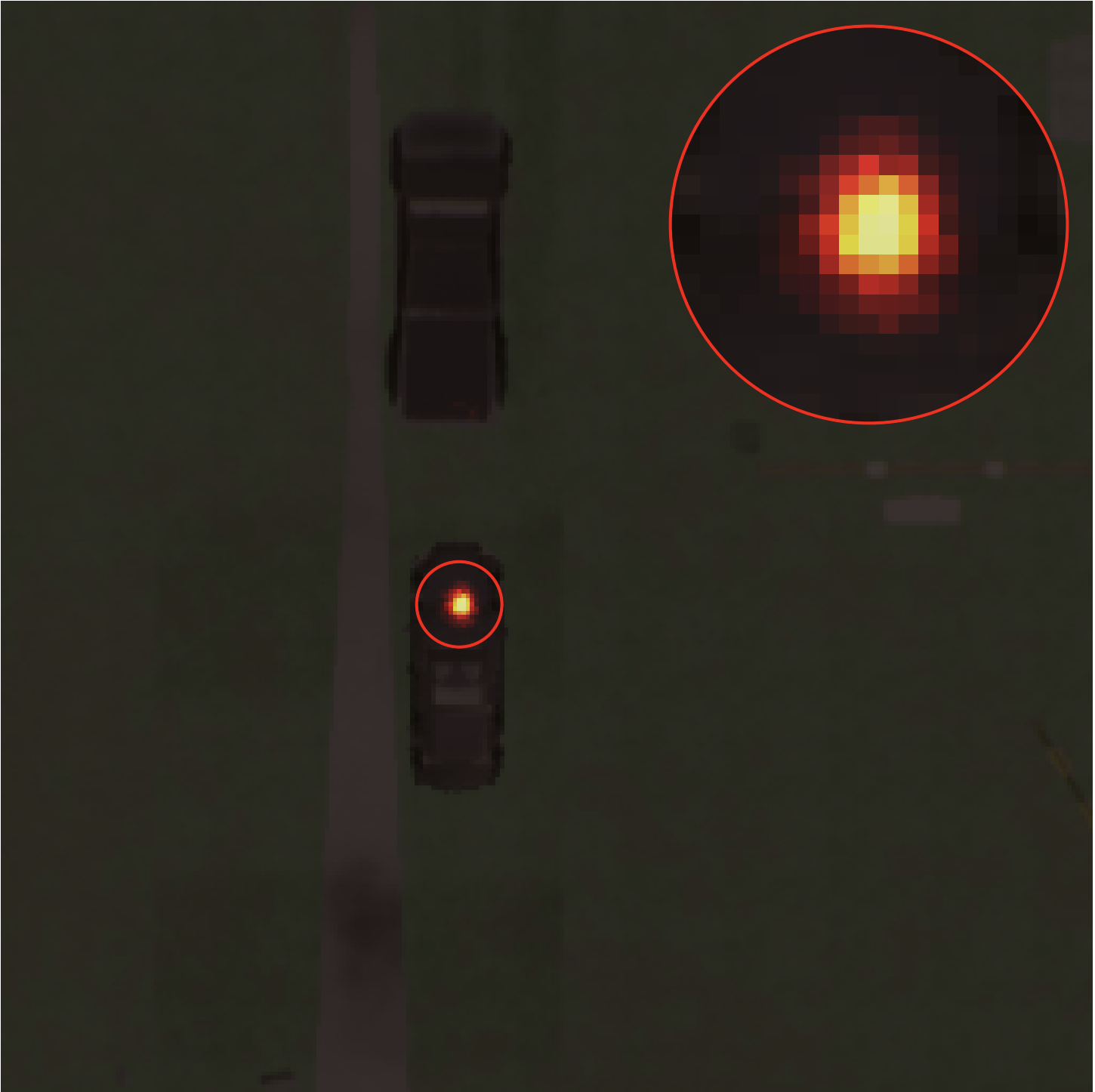}};
\draw (505,264) node  {\includegraphics[width=97.5pt,height=97.5pt]{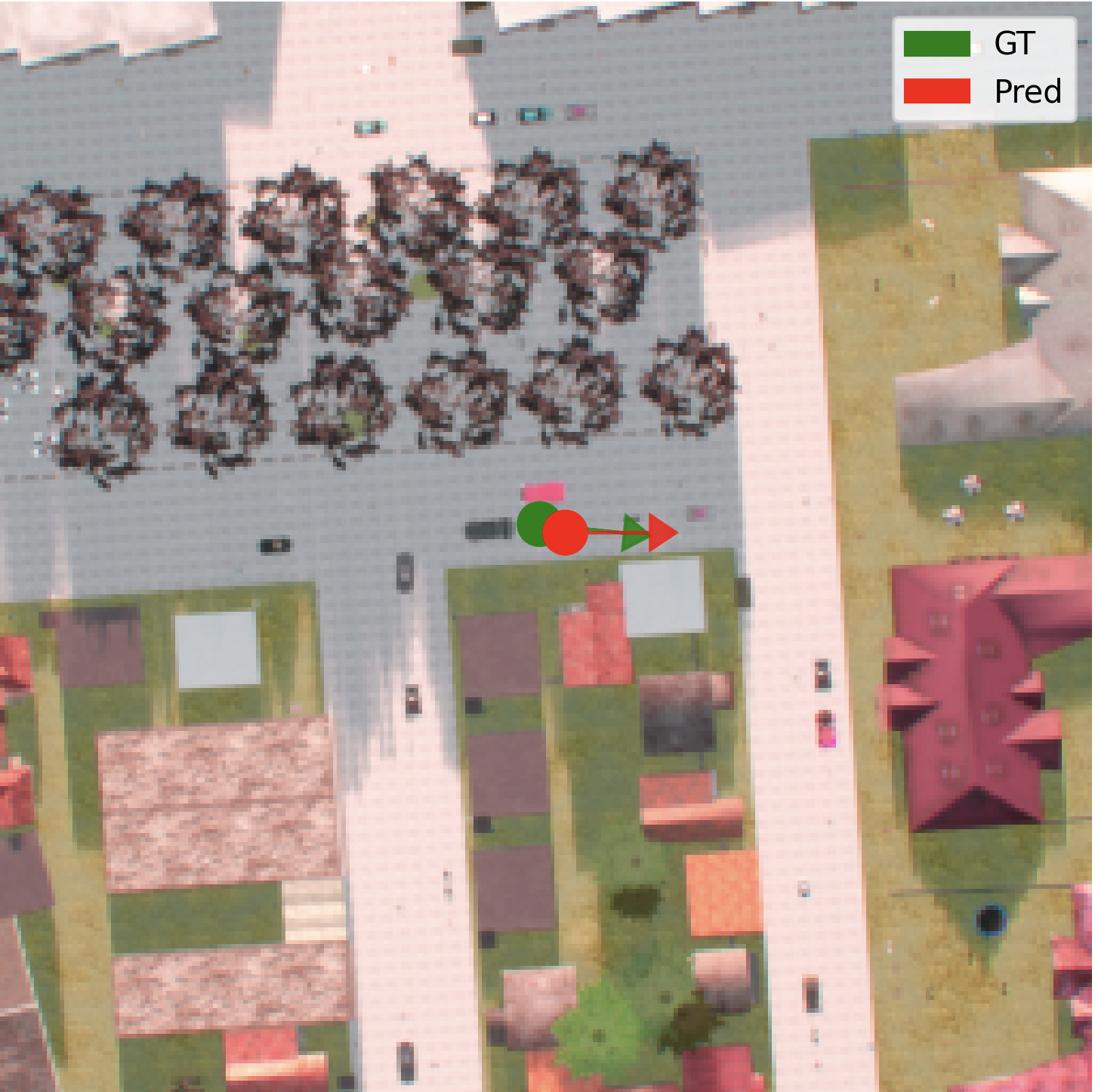}};
\draw (645,264) node  {\includegraphics[width=97.5pt,height=97.5pt]{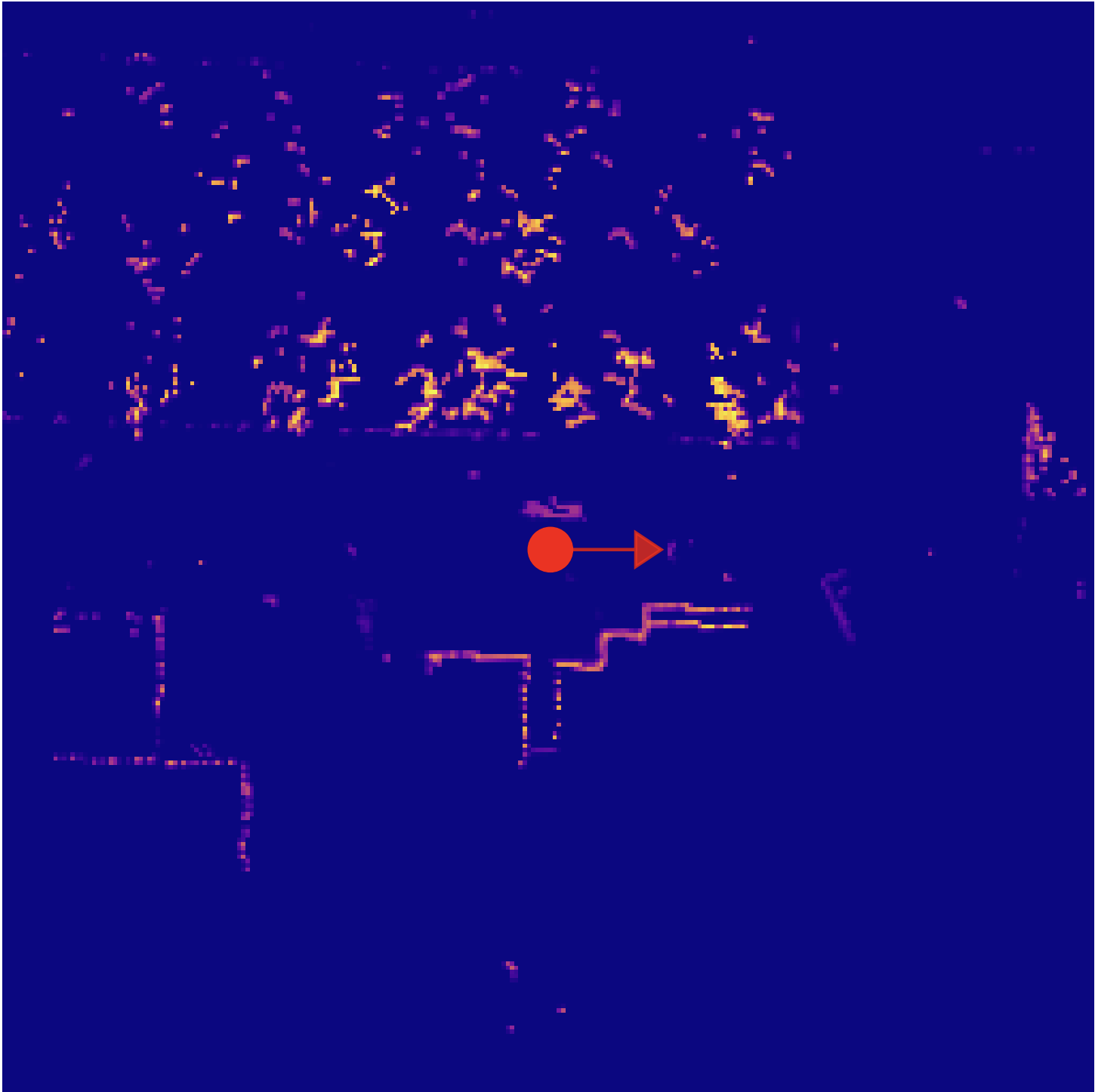}};
\draw (785,264) node  {\includegraphics[width=97.5pt,height=97.5pt]{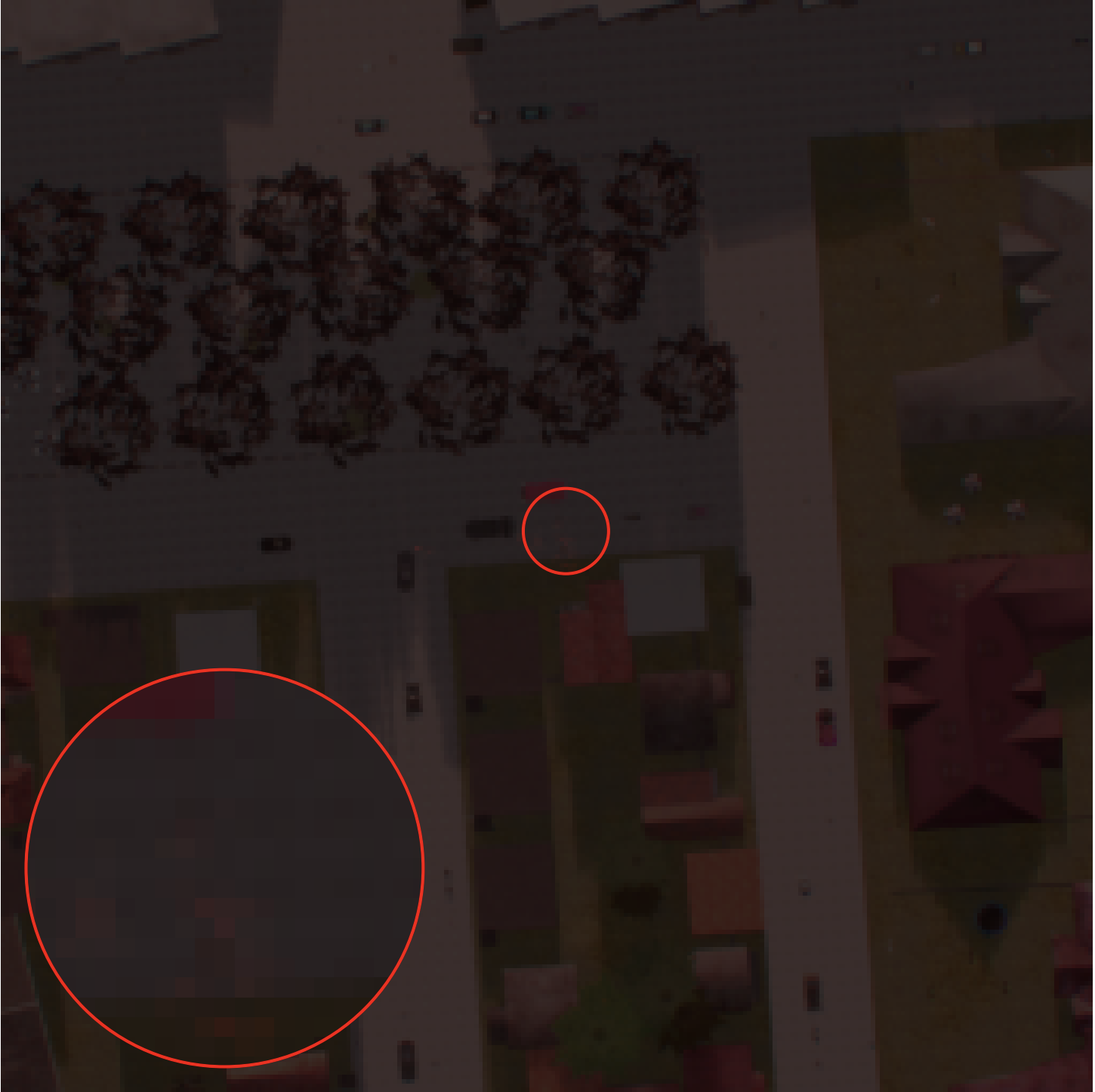}};

\draw (25,22.4) node [anchor=north west][inner sep=0.75pt]  [font=\footnotesize]  {$\Delta xy: 0.05m,\ \Delta \theta :\ 0.04^{\circ }$};
\draw (193,22.4) node [anchor=north west][inner sep=0.75pt]  [font=\footnotesize]  {Projected BEV};
\draw (325,22.4) node [anchor=north west][inner sep=0.75pt]  [font=\footnotesize]  {Location heatmap};
\draw (445,22.4) node [anchor=north west][inner sep=0.75pt]  [font=\footnotesize]  {$\Delta xy: 0.15m,\ \Delta \theta :\ 0.05^{\circ }$};
\draw (613,22.4) node [anchor=north west][inner sep=0.75pt]  [font=\footnotesize]  {Projected BEV};
\draw (745,22.4) node [anchor=north west][inner sep=0.75pt]  [font=\footnotesize]  {Location heatmap};
\draw (25,182.4) node [anchor=north west][inner sep=0.75pt]  [font=\footnotesize]  {$\Delta xy: 5.74m,\ \Delta \theta :\ 1.2^{\circ }$};
\draw (193,183.4) node [anchor=north west][inner sep=0.75pt]  [font=\footnotesize]  {Projected BEV};
\draw (325,182.4) node [anchor=north west][inner sep=0.75pt]  [font=\footnotesize]  {Location heatmap};
\draw (445,182.4) node [anchor=north west][inner sep=0.75pt]  [font=\footnotesize]  {$\Delta xy: 0.92m,\ \Delta \theta :\ 1.98^{\circ }$};
\draw (613,183.4) node [anchor=north west][inner sep=0.75pt]  [font=\footnotesize]  {Projected BEV};
\draw (745,182.4) node [anchor=north west][inner sep=0.75pt]  [font=\footnotesize]  {Location heatmap};

\end{tikzpicture}
}
\caption{\small Qualitative localization results on the CARLA dataset. Each example shows an aerial image with predicted pose, the projected BEV, and a heatmap of vehicle position likelihood. The synthetic environment provides aligned aerial-ground pairs for precise evaluation, achieving consistent sub-meter and sub-degree accuracy in various conditions.}
\label{fig:vis_carla}
\end{figure*}
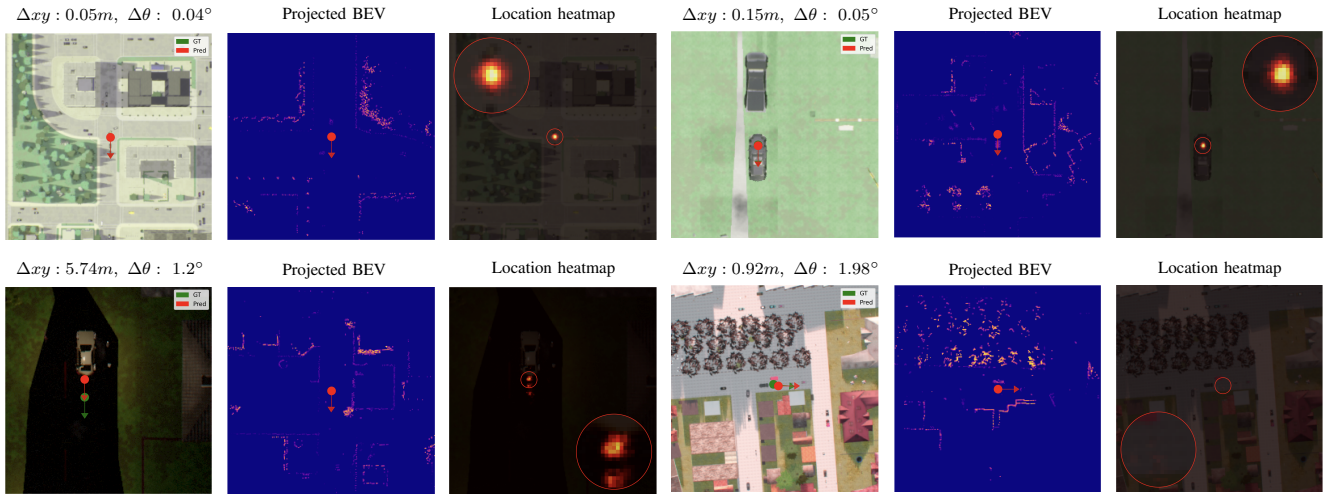
\subsection{Cross-view localization with aerial and ground data}
Cross-view localization matches ground-level images with geo-referenced aerial databases under extreme viewpoint differences. CVM-Net \cite{CVM-Net} introduced deep Siamese networks with NetVLAD \cite{netvlad} descriptors and metric learning, while later works added rotation awareness and improved local feature matching \cite{shi2020optimal,liu2019lending}. Recent methods integrate spatially aware correlation and multi-scale descriptors \cite{patch-netvlad} for improved robustness.

Beyond image-only approaches, several works align LiDAR or BEV representations with aerial maps. BEVPlace++ \cite{luo2025bevplace++} uses BEV images and CNNs with rotation equivariance for 3-DoF pose estimation. AGL-Net \cite{guan2024agl} aligns LiDAR point clouds with aerial maps using a two-stage matching network with scale and skeleton loss for robust, scale-invariant pose estimation.

\subsection{Cross-modal attention and BEV-centric fusion}

Transformer-based cross-attention \cite{vaswani2017attention} has become standard for fusing heterogeneous views into BEV representations. BEVFormer \cite{li2024bevformer} learns BEV queries with spatial cross-attention to camera features and temporal
self-attention, while Lift-Splat-Shoot \cite{philion2020lift} pioneered
differentiable lifting of perspective images into BEV. BEVFusion \cite{bevfusion} extends this paradigm by fusing multi-modal features from cameras and LiDAR in a unified BEV space, while MapTR \cite{map-tr} focuses on online mapping and localization using BEV representations. End-to-end driving systems such as TransFuser \cite{chitta2022transfuser} couple LiDAR and camera streams with global attention.

\section{METHOD}

We propose \ours, a cross-modal attention framework for aerial-ground vehicle localization that integrates LiDAR point clouds with aerial imagery through feature extraction, cross-modal attention, and contrastive learning.

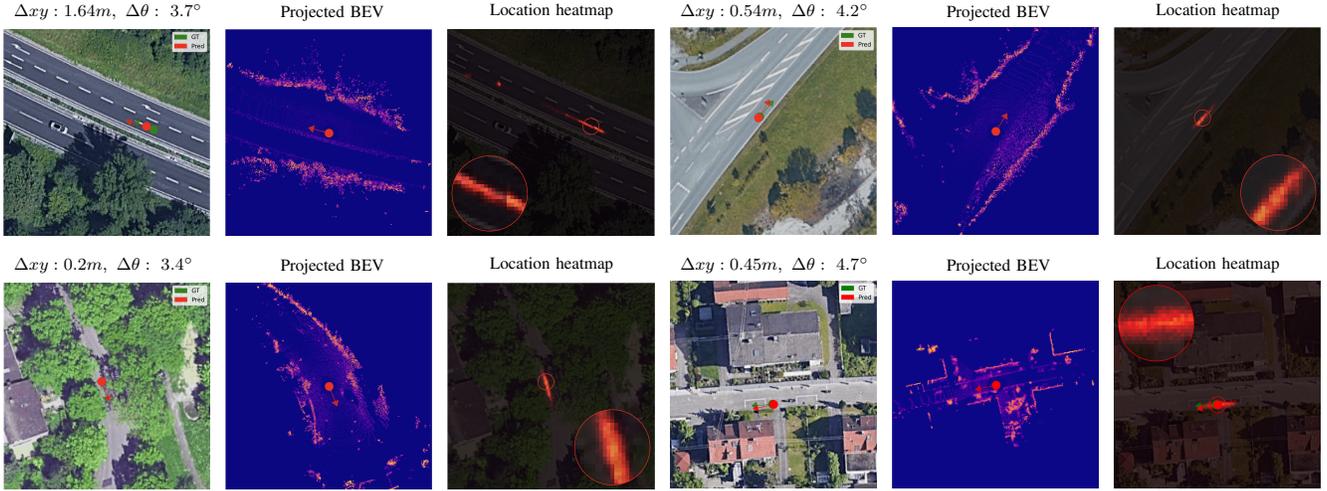
\begin{figure*}[htbp]
\centering
\tikzset{every picture/.style={line width=0.75pt}} 

\scalebox{0.8}{
\begin{tikzpicture}[x=0.75pt,y=0.75pt,yscale=-1,xscale=1]

\draw (85,105) node  {\includegraphics[width=97.5pt,height=97.5pt]{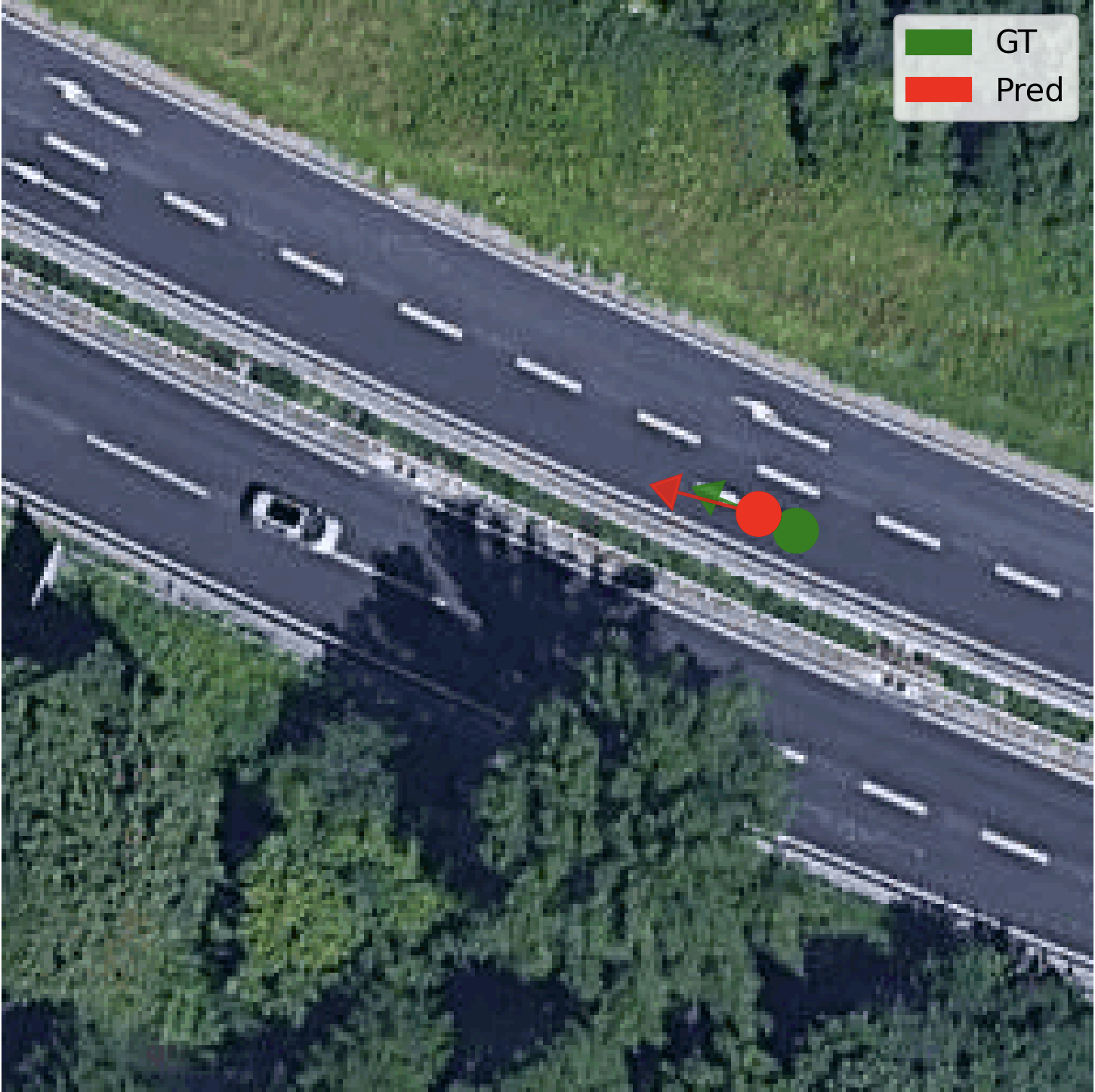}};
\draw (225,105) node  {\includegraphics[width=97.5pt,height=97.5pt]{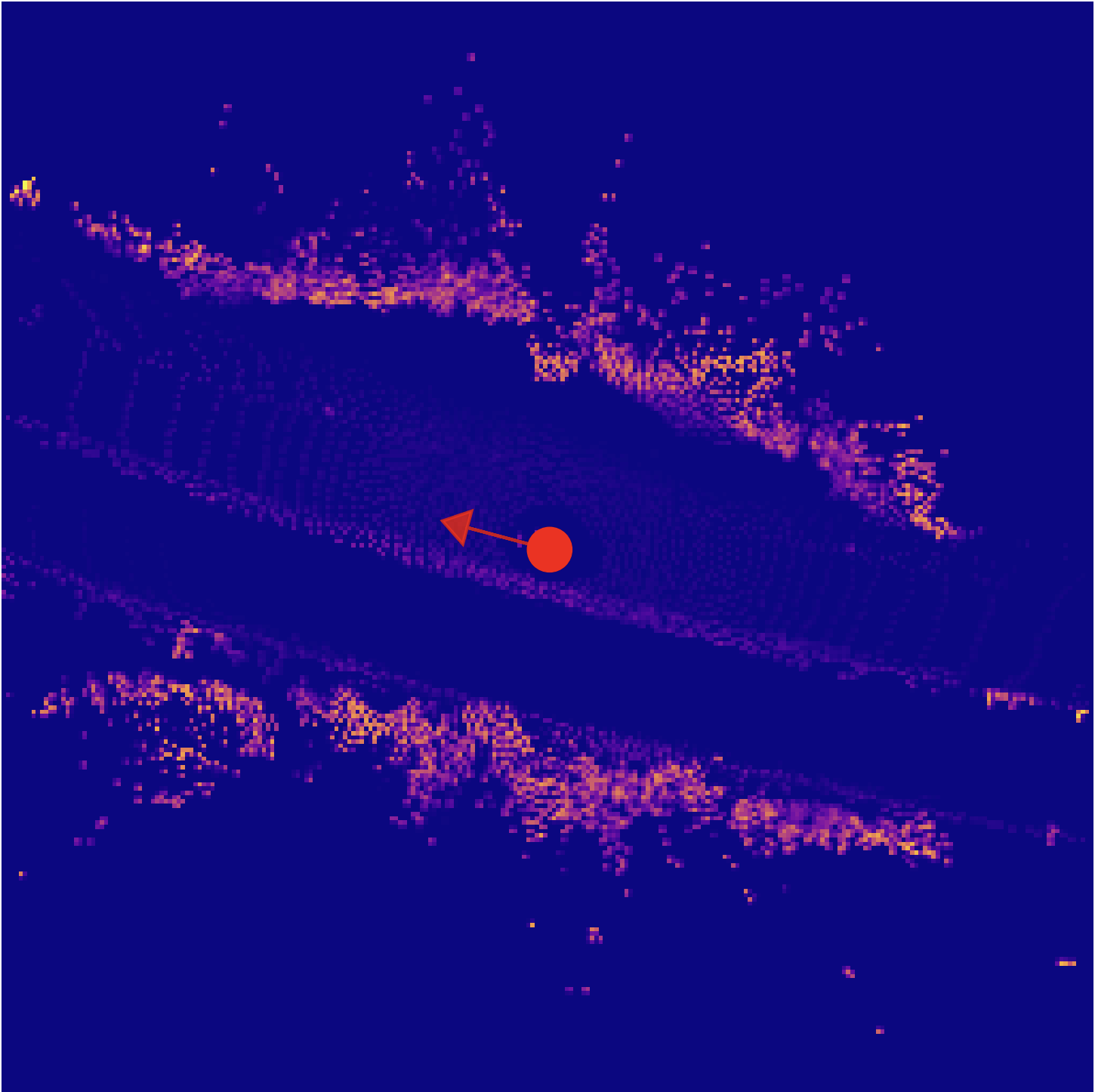}};
\draw (365,105) node  {\includegraphics[width=97.5pt,height=97.5pt]{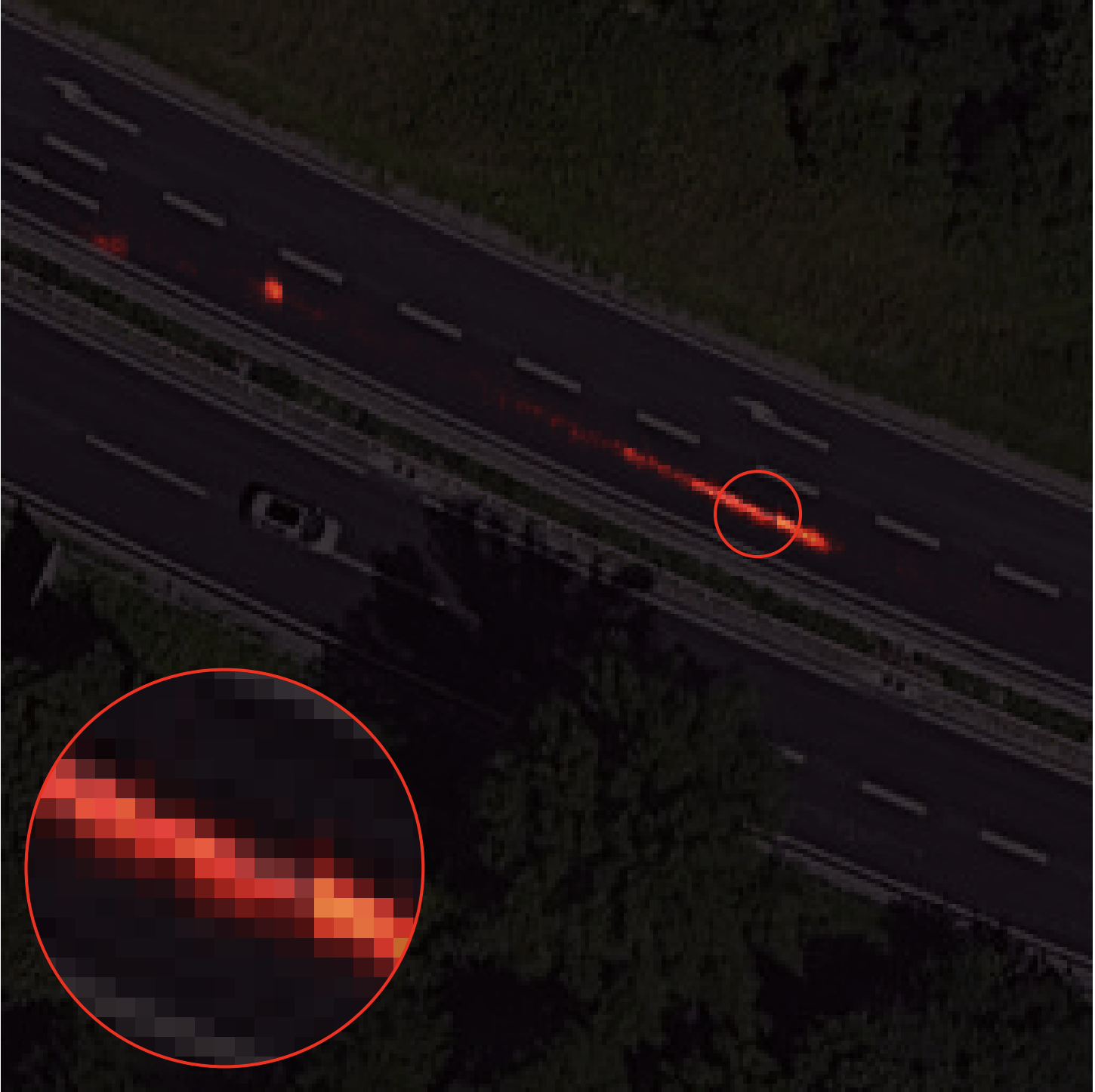}};
\draw (85,265) node  {\includegraphics[width=97.5pt,height=97.5pt]{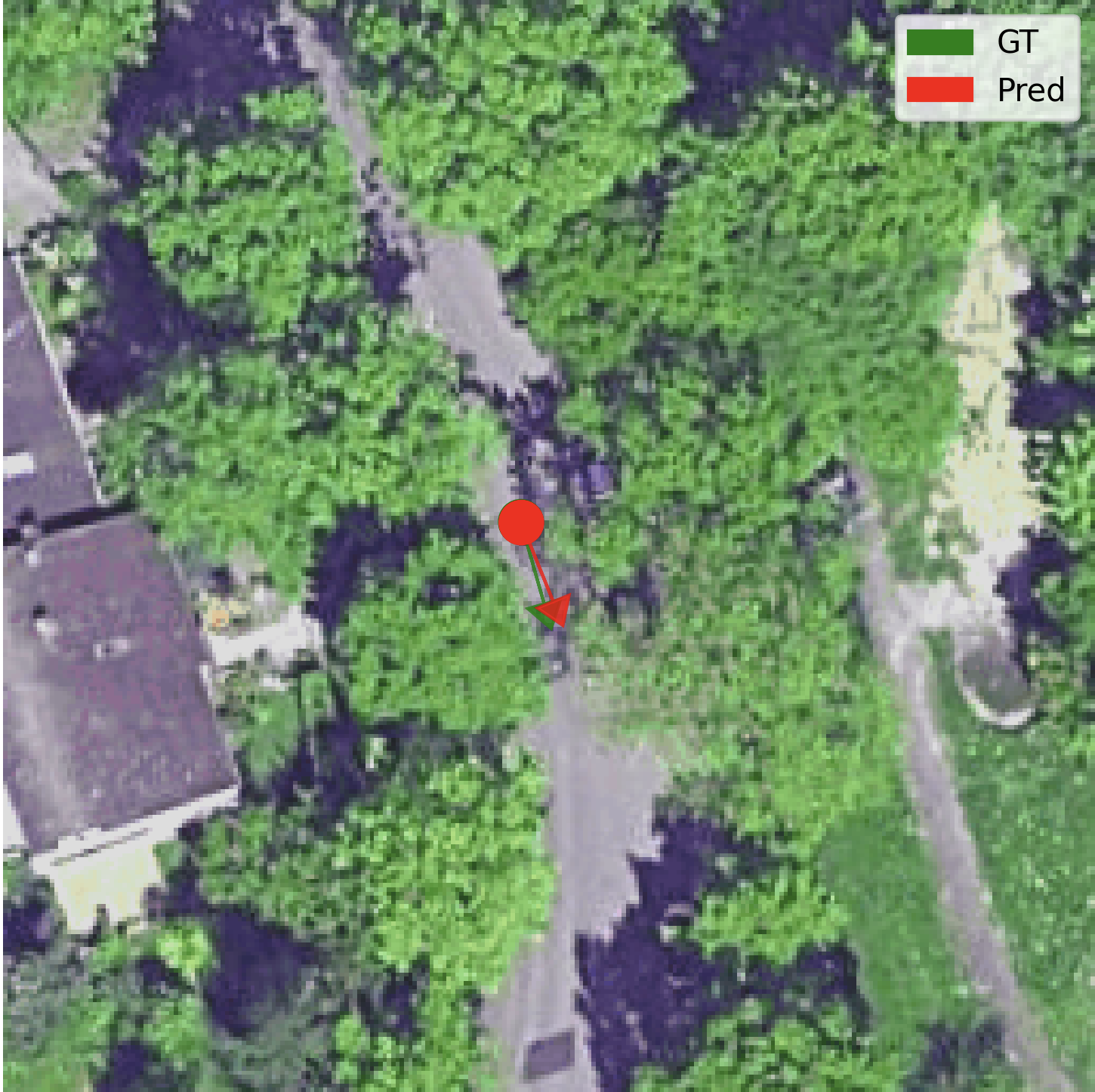}};
\draw (225,265) node  {\includegraphics[width=97.5pt,height=97.5pt]{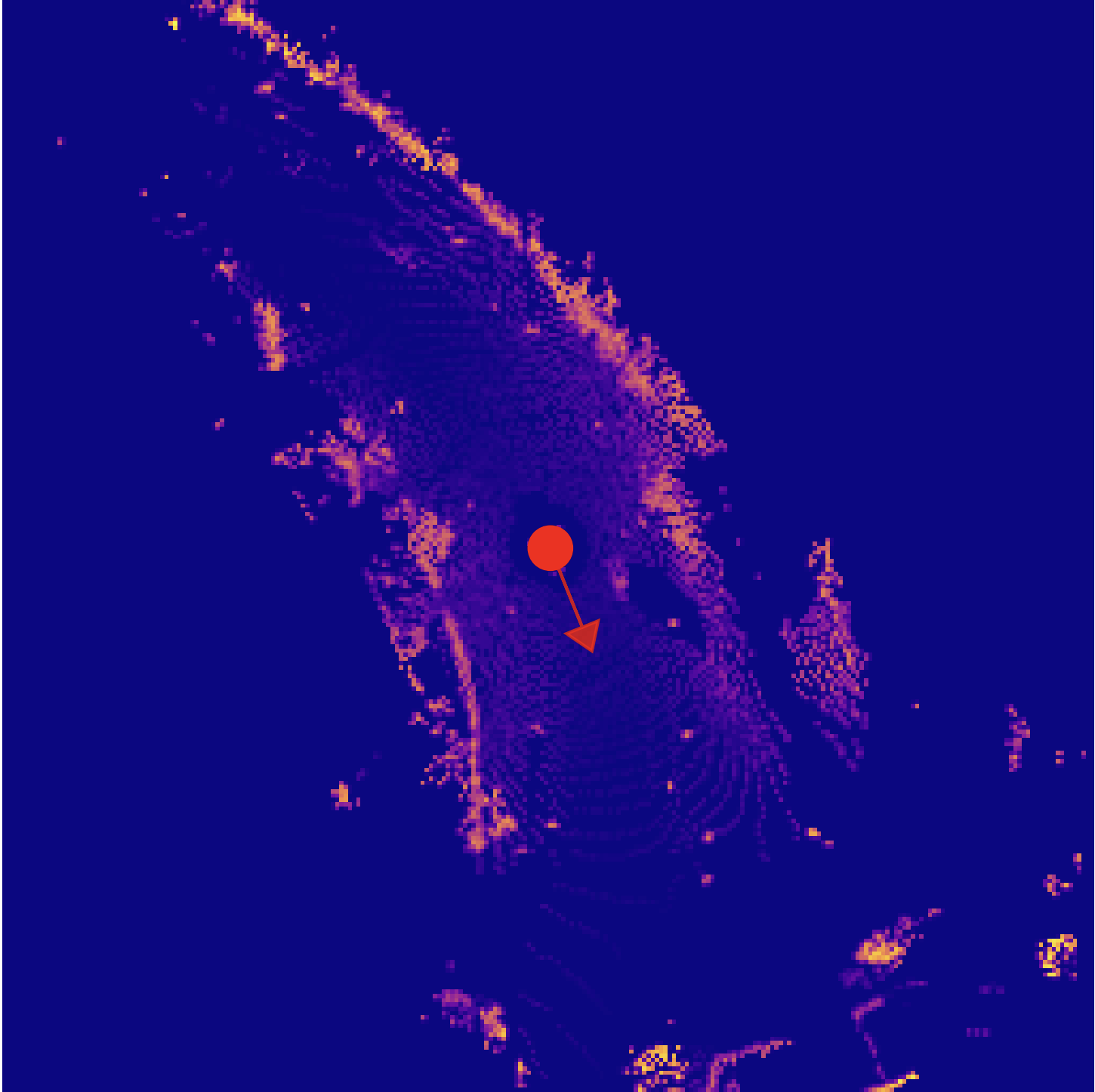}};
\draw (365,265) node  {\includegraphics[width=97.5pt,height=97.5pt]{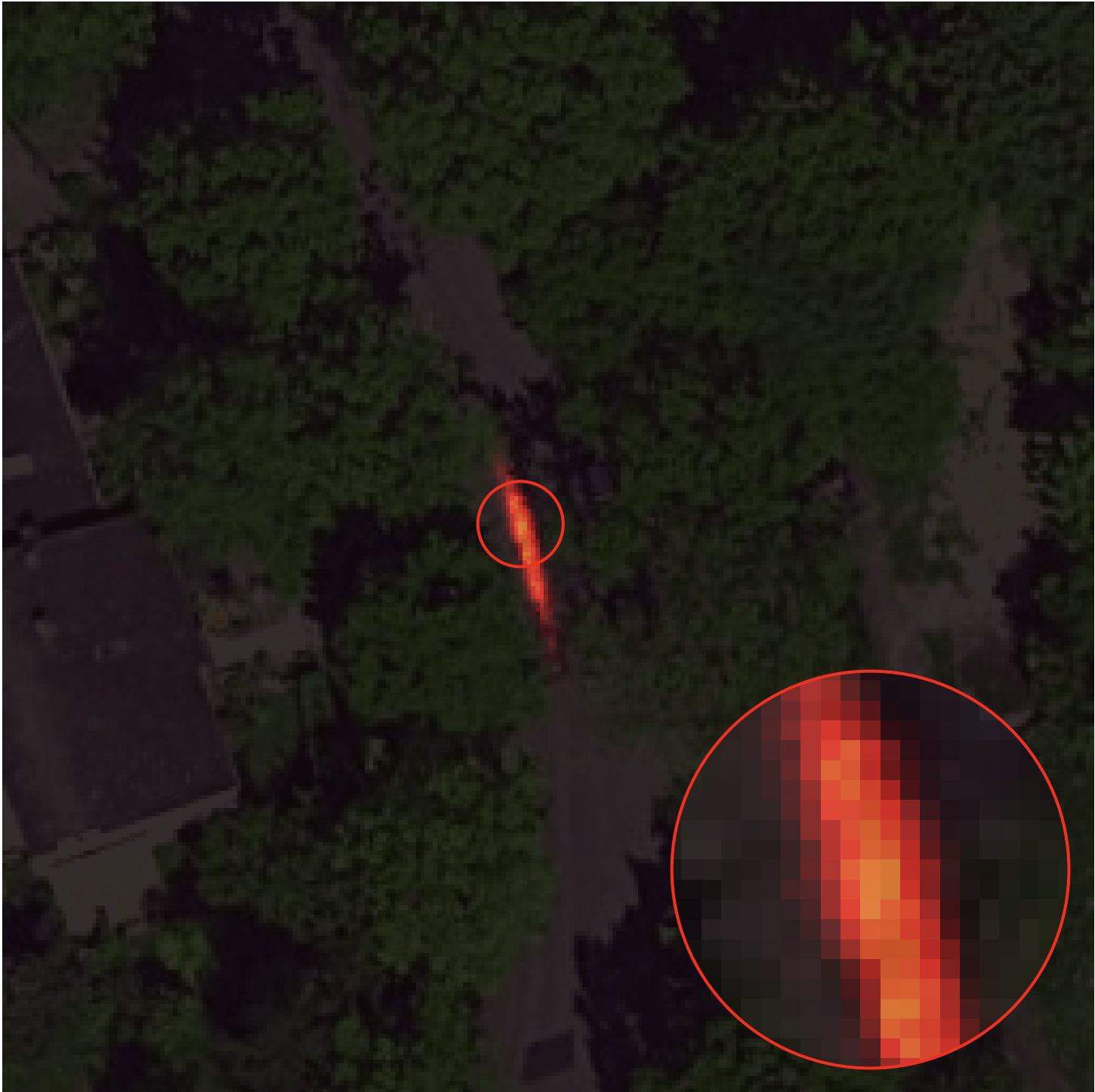}};
\draw (505,104) node  {\includegraphics[width=97.5pt,height=97.5pt]{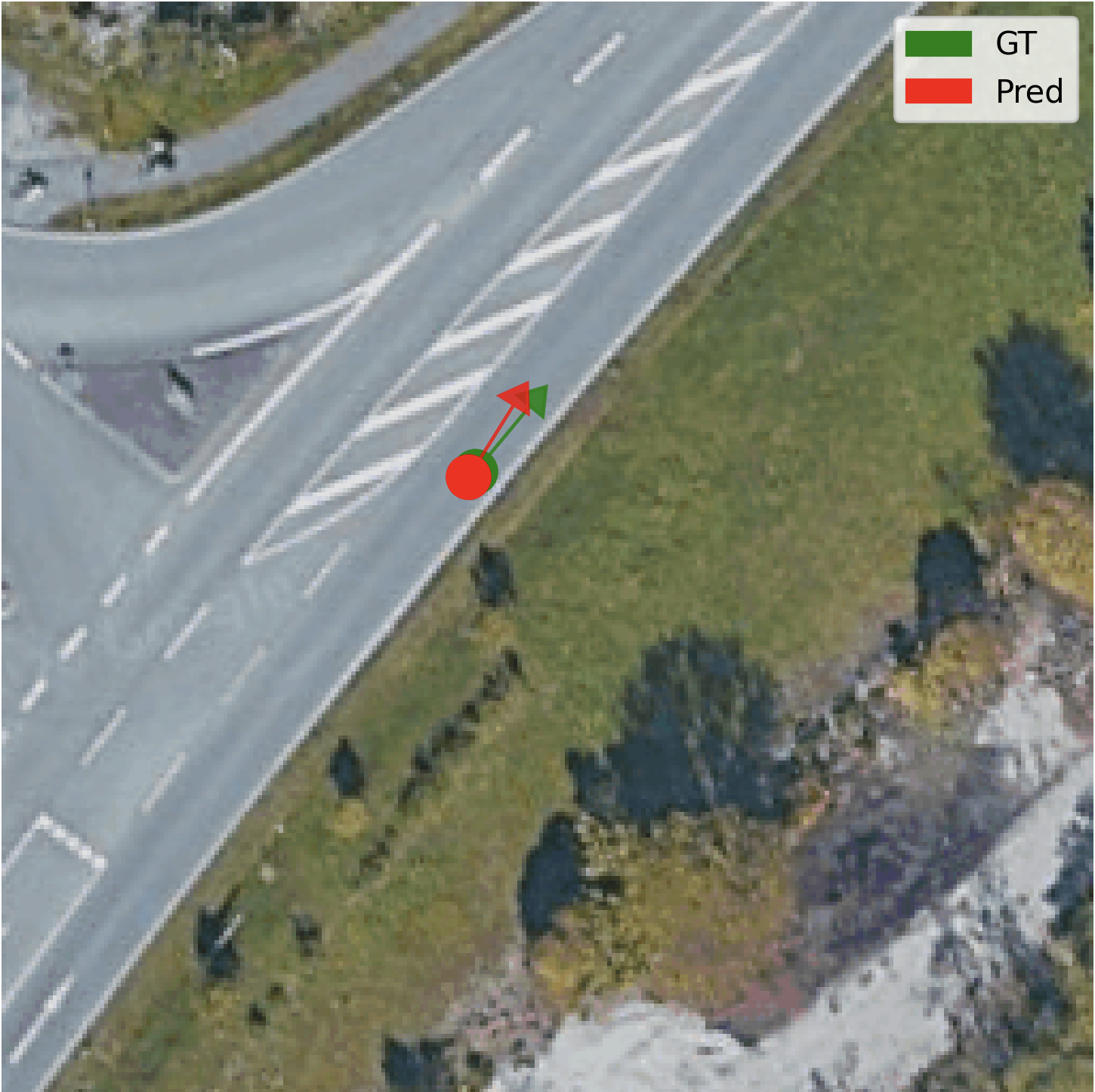}};
\draw (645,104) node  {\includegraphics[width=97.5pt,height=97.5pt]{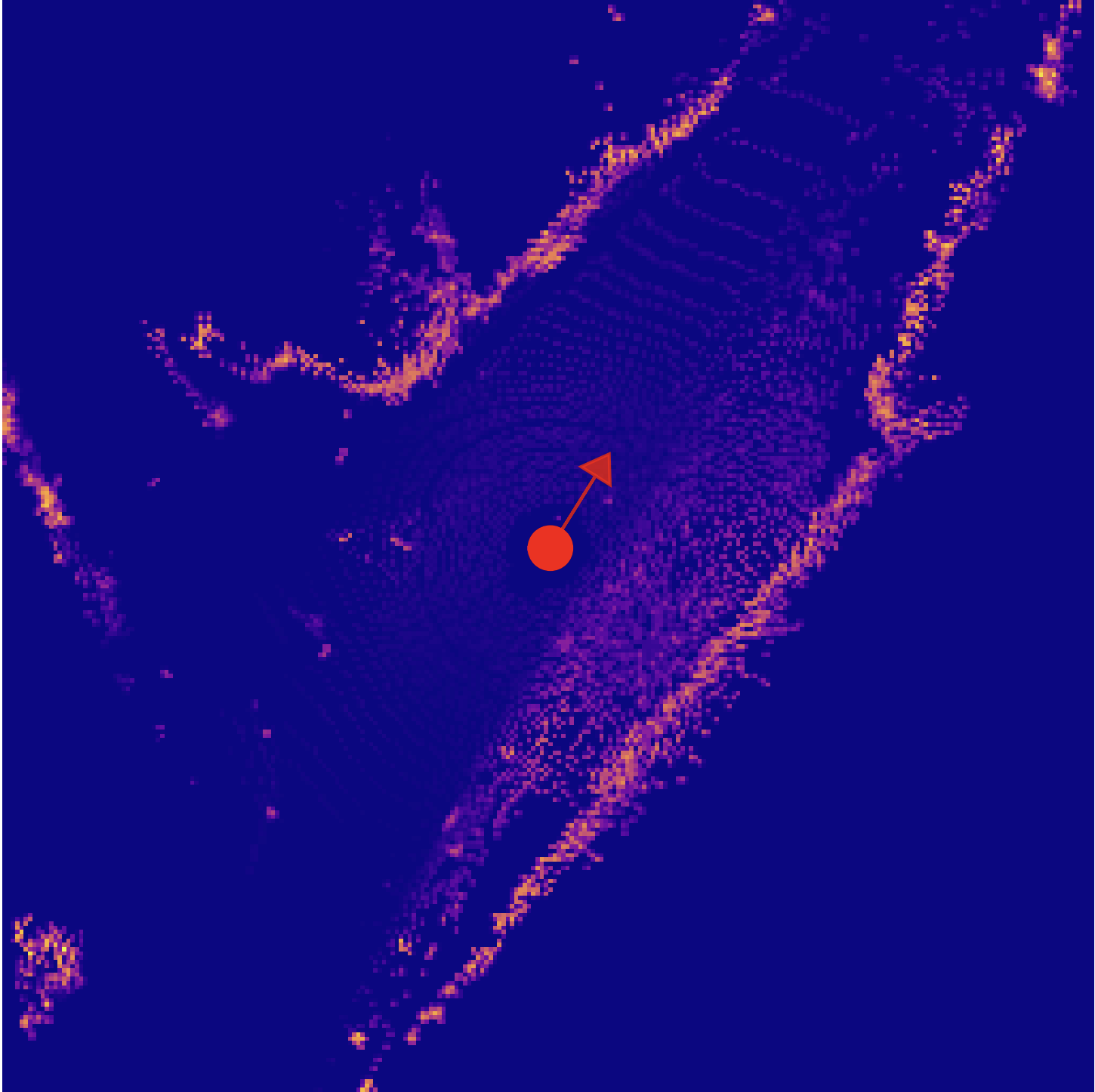}};
\draw (785,104) node  {\includegraphics[width=97.5pt,height=97.5pt]{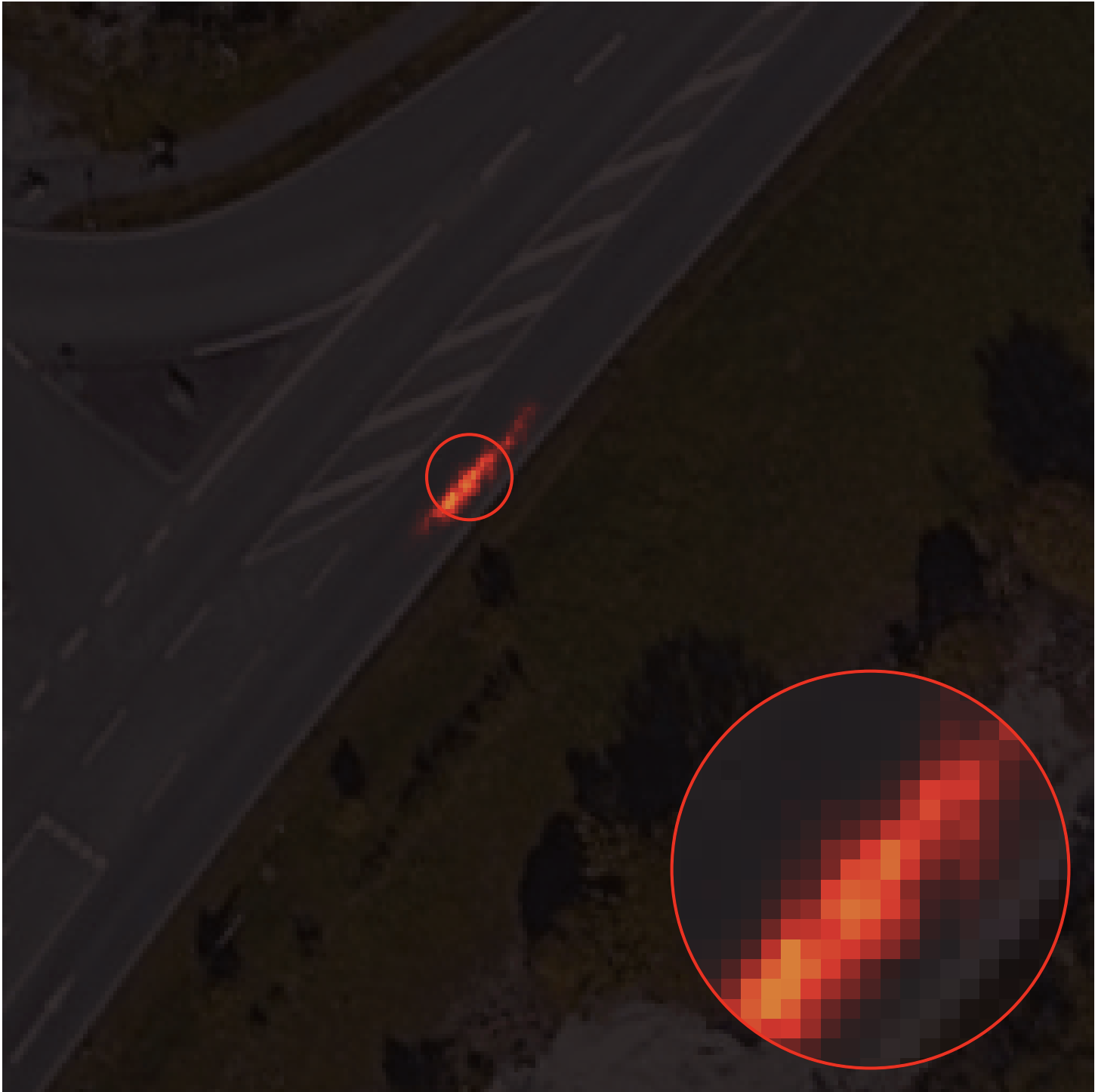}};
\draw (505,264) node  {\includegraphics[width=97.5pt,height=97.5pt]{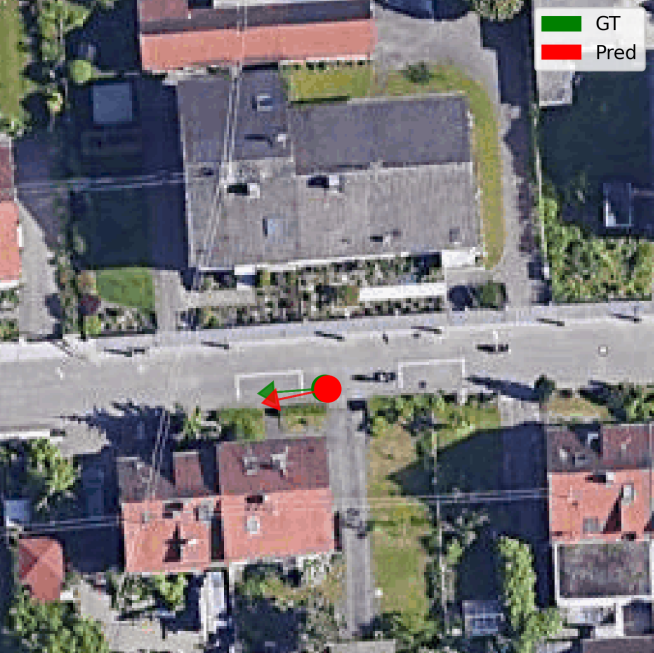}};
\draw (645,264) node  {\includegraphics[width=97.5pt,height=97.5pt]{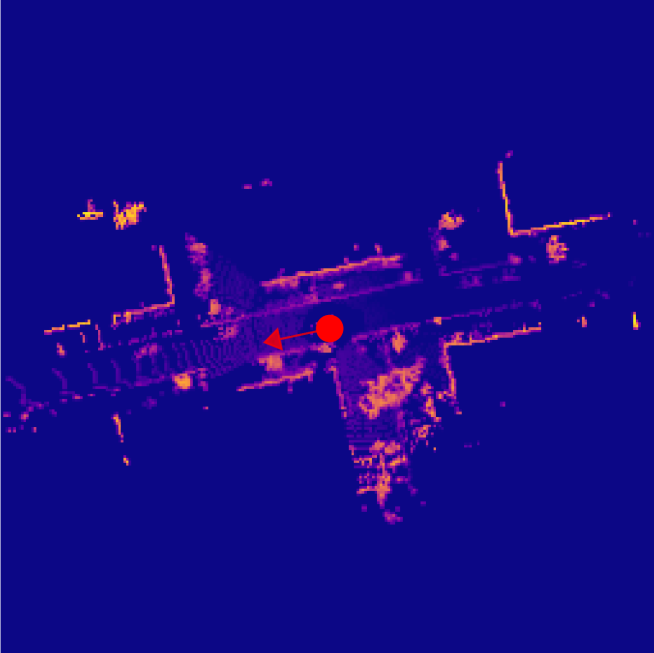}};
\draw (785,264) node  {\includegraphics[width=97.5pt,height=97.5pt]{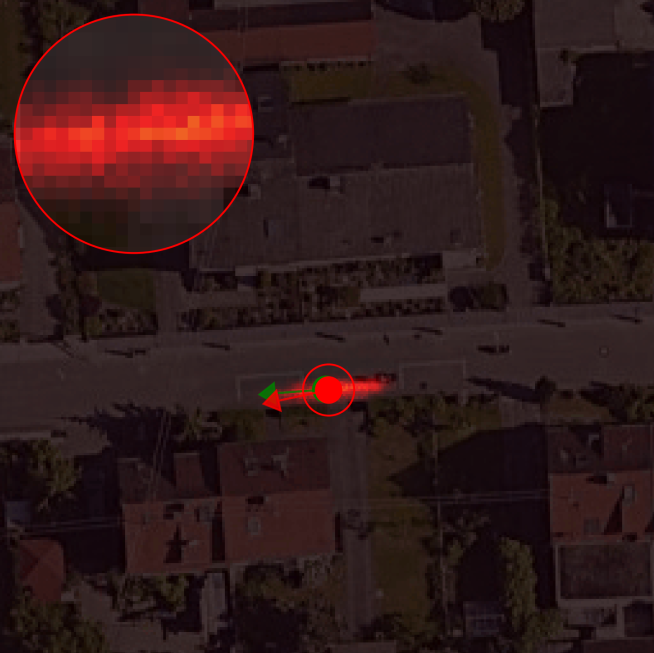}};

\draw (25,22.4) node [anchor=north west][inner sep=0.75pt]  [font=\footnotesize]  {$\Delta xy: 1.64m,\ \Delta \theta :\ 3.7^{\circ }$};
\draw (193,23.4) node [anchor=north west][inner sep=0.75pt]  [font=\footnotesize]  {Projected BEV};
\draw (325,22.4) node [anchor=north west][inner sep=0.75pt]  [font=\footnotesize]  {Location heatmap};
\draw (445,22.4) node [anchor=north west][inner sep=0.75pt]  [font=\footnotesize]  {$\Delta xy: 0.54m,\ \Delta \theta :\ 4.2^{\circ }$};
\draw (613,23.4) node [anchor=north west][inner sep=0.75pt]  [font=\footnotesize]  {Projected BEV};
\draw (745,22.4) node [anchor=north west][inner sep=0.75pt]  [font=\footnotesize]  {Location heatmap};
\draw (25,182.4) node [anchor=north west][inner sep=0.75pt]  [font=\footnotesize]  {$\Delta xy: 0.2m,\ \Delta \theta :\ 3.4^{\circ }$};
\draw (193,183.4) node [anchor=north west][inner sep=0.75pt]  [font=\footnotesize]  {Projected BEV};
\draw (325,182.4) node [anchor=north west][inner sep=0.75pt]  [font=\footnotesize]  {Location heatmap};
\draw (445,182.4) node [anchor=north west][inner sep=0.75pt]  [font=\footnotesize]  {$\Delta xy: 0.45m,\ \Delta \theta :\ 4.7^{\circ }$};
\draw (613,183.4) node [anchor=north west][inner sep=0.75pt]  [font=\footnotesize]  {Projected BEV};
\draw (745,182.4) node [anchor=north west][inner sep=0.75pt]  [font=\footnotesize]  {Location heatmap};

\end{tikzpicture}
}
\caption{\small Qualitative localization results on the KITTI dataset. Despite challenges from sensor noise and temporal misalignment, \ours produces accurate localization with errors typically within a few meters and degrees.}
\label{fig:vis_kitti}
\end{figure*}
\subsection{Problem Formulation}

Given a local LiDAR point cloud $\mathcal{P} = \{p_i\}_{i=1}^N$ where $p_i \in \mathbb{R}^3$ represents 3D coordinates, and a corresponding aerial image $\mathcal{I} \in \mathbb{R}^{H \times W \times 3}$, our goal is to estimate the vehicle's pose $\mathbf{T} = (x, y, \theta)$ in the aerial map coordinate system, where $(x, y)$ denotes the 2D position and $\theta$ represents the heading angle. 

The key challenge lies in bridging the viewpoint gap between ground-level LiDAR observations and overhead aerial imagery. We project the LiDAR point cloud to a BEV representation $\mathcal{B} \in \mathbb{R}^{H' \times W' \times 1}$, which naturally aligns with the aerial perspective and preserves geometric structure essential for localization.

\subsection{Network Architecture}

An overview of the proposed architecture is shown in Fig. \ref{fig:architecture}. \ours comprises modality-specific encoders, a cross-modal attention module, and a likelihood map decoder, complemented by a contrastive learning objective for robust feature alignment.

\subsubsection{Feature Extraction}
We employ separate encoders for the aerial map and the BEV representation of the ground LiDAR point cloud. This design choice allows each encoder to specialize in processing its respective modality while maintaining architectural consistency for effective cross-modal fusion.

\textbf{BEV Encoder:} Our BEV encoder processes the projected point cloud data using a ResNet-34 backbone with multi-scale feature fusion. The encoder extracts hierarchical features at three scales ($1/8$, $1/16$, $1/32$) and fuses them through a refinement network with spatial and channel attention mechanisms. 

The spatial attention mechanism highlights geometrically salient regions, while the channel attention mechanism emphasizes discriminative feature dimensions. To handle viewpoint variations, particularly rotation differences between LiDAR scans and aerial imagery, we incorporate a Fourier Transform layer that provides rotation-invariant representations. The magnitude spectrum of the 2D Fourier transform is invariant to rotations, as rotating an image corresponds to a phase shift in the frequency domain, leaving the magnitude spectrum unchanged.

This rotation invariance is particularly useful for BEV representations where the vehicle's heading angle may vary significantly relative to the aerial map orientation. The Fourier transform captures global structural patterns while being robust to rotational misalignment:

\vspace{-10pt}
\begin{equation}
\mathcal{F}_{bev} = \text{FFT}(\text{Attention}(\text{Refine}(\text{MultiScale}(\mathcal{B}))))
\end{equation}

where $\text{MultiScale}(\cdot)$ extracts multi-scale features, $\text{Refine}(\cdot)$ applies feature refinement with attention, and $\text{FFT}(\cdot)$ computes the log-amplitude spectrum of the Fourier transform.

\textbf{Aerial Encoder:} The aerial image encoder also utilizes the same ResNet-34 backbone to extract spatial features from the overhead imagery. This architectural consistency ensures that both encoders produce features in a compatible representation space, facilitating effective cross-modal attention. The encoder processes the aerial image $\mathcal{I}$ to produce feature maps $\mathcal{F}_{aerial} \in \mathbb{R}^{H/32 \times W/32 \times D}$ where $D$ is the feature dimension. The aerial encoder captures semantic information such as road networks, building layouts, and vegetation patterns that aid localization.

\subsubsection{Cross-Modal Attention}

To bridge the modality gap between BEV and aerial features, we introduce a cross-modal attention mechanism that enables dynamic feature alignment. Traditional fusion approaches, such as simple concatenation or element-wise operations, fail to capture the complex spatial correspondences between different viewpoints. Our attention mechanism addresses this by allowing each modality to selectively attend to relevant regions in the other modality.

This bidirectional attention enables mutual information exchange between aerial and BEV features. Each modality can selectively focus on relevant regions in the other modality. BEV features leverage aerial cues to locate corresponding structures such as roads or buildings, while aerial features exploit BEV geometry to validate ground-level details. The attention computation follows the standard transformer formulation \cite{vaswani2017attention}:

\vspace{-7pt}
\begin{align}
\text{Attention}_{a2b} &= \text{Softmax}\left(\frac{Q_a K_b^T}{\sqrt{d_k}} + R_{pos}\right) V_b \\
\text{Attention}_{b2a} &= \text{Softmax}\left(\frac{Q_b K_a^T}{\sqrt{d_k}} + R_{pos}\right) V_a
\end{align}

where $Q_a, K_a, V_a$ and $Q_b, K_b, V_b$ are query, key, and value projections for aerial and BEV features respectively, $d_k$ is the dimension of the key vectors, and $R_{pos}$ represents learnable relative positional encoding. The bidirectional design ensures that both modalities can influence each other, with aerial features providing global context to local BEV observations.

The relative positional encoding $R_{pos}$ is crucial for maintaining spatial awareness across different scales and viewpoints. The attended features are then fused through a multi-scale fusion network that preserves both local and global information:
\vspace{-10pt}
\begin{equation}
\mathcal{F}_{fused} = \text{Fusion}(\text{Concat}(\mathcal{F}_{aerial}, \text{Attention}_{a2b}(\mathcal{F}_{bev})))
\end{equation}

\subsubsection{Likelihood Map Decoder}

The likelihood map decoder generates spatial probability distributions for both location and orientation estimation. Unlike direct regression approaches that output point estimates, our probabilistic formulation provides uncertainty quantification and enables robust handling of ambiguous scenarios (e.g., symmetric intersections, repetitive building patterns). The decoder employs a U-Net style architecture with skip connections to preserve fine-grained spatial details while incorporating global context:

\begin{itemize}
\item \textbf{Location Map:} $\mathcal{M}_{loc} \in \mathbb{R}^{H \times W}$ representing the probability distribution over possible vehicle locations. This probabilistic representation naturally handles localization ambiguity and provides confidence measures.
\item \textbf{Orientation Map:} $\mathcal{M}_{ori} \in \mathbb{R}^{K \times H \times W}$ where $K$ is the number of orientation bins, encoding orientation probabilities at each spatial location. The discrete binning approach simplifies the orientation estimation task while maintaining sufficient angular resolution.
\item \textbf{Confidence Score:} $c \in [0,1]$ indicating the reliability of the prediction, learned from the accuracy of the predicted pose relative to ground truth.
\end{itemize}

The final pose estimate is obtained by finding the maximum likelihood location and orientation:

\vspace{-10pt}
\begin{align}
(x^*, y^*) &= \arg\max_{(x,y)} \mathcal{M}_{loc}(x,y) \\
\theta^* &= \arg\max_k \mathcal{M}_{ori}(k, x^*, y^*)
\end{align}

This approach provides interpretable uncertainty estimates and enables downstream systems to make informed decisions based on localization confidence.

\begin{table*}[!htp]
    \begin{center}
    \resizebox{\textwidth}{!}{
        \setlength{\tabcolsep}{4pt}
        \begin{tabular}{cccccccccccc}
            \toprule
            \multirow{2}[2]{*}{
            \begin{tabular}{c}
                \textbf{Method}\\
            \end{tabular}
            } &     
            \multirow{2}[2]{*}{
            \begin{tabular}{c}
                \textbf{Map Size} \\ (in pixels)
            \end{tabular}
            } & 
            \multirow{2}[2]{*}{
            \begin{tabular}{c}
                {\textbf{Avg. Loc. / Ori.}} \\ {\textbf{Error} (m / \degree) $\downarrow$}
            \end{tabular}
            } & \multicolumn{3}{c}{\textbf{Lat. R@Xm} $\uparrow$} &  \multicolumn{3}{c}{\textbf{Long. R@Xm} $\uparrow$} &  \multicolumn{3}{c}{{ \textbf{Ori. R@X\degree $\uparrow$}
            }}  \\ \cmidrule(lr){4-6} \cmidrule(lr){7-9}  \cmidrule(lr){10-12}              
            &  &  & 1m & 3m & 5m  & 1m & 3m & 5m & 1\degree & 3\degree & 5\degree \\
             \midrule
            OrienterNet & $256\times 256$ & 2.23 / 19.15 & 66.87 & 87.67 & 98.61 & 44.38 & 81.51 & 97.69 & 55.78 & 62.71 & 66.26\\
            AGL-Net & $256\times 256$ & 1.83 / 3.76 & 76.58 & 92.14 & 100.0 & 44.07 & 88.44 & 99.54 & 57.47 & 68.41 & 74.42\\
            \ours & $256\times 256$ & \first{1.57} / \first{3.22} & \first{78.12} & \first{95.38} & \first{100.0} & \first{47.83} & \first{91.37} & \first{100.0} & \first{57.92} & \first{70.86} & \first{78.95}\\
             \midrule
            OrienterNet & $1024\times 1024$ & 18.75 / 82.71 & 33.44 & 40.99 & 56.55 & 13.41 & 18.64 & 22.50 & 30.35 & 31.43 & 32.51\\
            AGL-Net & $1024\times 1024$ & 14.96 / 57.25 & 50.54 & 57.94 & 69.49 & 18.64 & 24.19 & 29.89 & 42.06 & 44.22 & 46.22\\
            \textbf{\ours} & $1024\times 1024$ & \first{5.75} / \first{10.93} & \first{53.04} & \first{69.95} & \first{87.38} & \first{25.41} & \first{39.60} & \first{60.55} & \first{44.84} & \first{61.17} & \first{66.87}\\
            \bottomrule
        \end{tabular}
    }
    \end{center}
\caption{\small Results on the synthetic CARLA dataset. The reported numbers for AGL-Net and OrienterNet are taken directly from the AGL-Net paper. \ours consistently achieves more accurate and robust performance than the baselines across both resolutions.}
\label{tab:carla_results}
\vspace{-3mm}
\end{table*}

\subsection{Contrastive Learning Module}

To enhance feature discriminability and improve cross-modal alignment, we integrate a contrastive learning module that learns shared feature representations between aerial and BEV modalities. The contrastive learning strategy addresses the inherent differences between LiDAR and visual data by learning a shared embedding space where semantically similar aerial-BEV pairs are close together, while dissimilar pairs are pushed apart.

The module projects aerial and BEV features to a shared embedding space through separate projection heads:

\vspace{-10pt}
\begin{align}
\mathbf{z}_{aerial} &= \text{Project}_{aerial}(\text{GlobalPool}(\mathcal{F}_{aerial})) \\
\mathbf{z}_{bev} &= \text{Project}_{bev}(\text{GlobalPool}(\mathcal{F}_{bev}))
\end{align}

where $\text{GlobalPool}(\cdot)$ performs global average pooling to aggregate spatial features, and the projection heads consist of two-layer MLPs with ReLU activation and layer normalization. The projected embeddings are normalized to unit length for cosine similarity computation.

The contrastive loss uses the standard InfoNCE formulation to encourage positive pairs (matching aerial-BEV pairs) to have high similarity while pushing negative pairs apart:

\vspace{-10pt}
\begin{equation}
\mathcal{L}_{contrastive} = -\log \frac{\exp(\mathbf{z}_{aerial}^T \mathbf{z}_{bev} / \tau)}{\sum_{j=1}^{B} \exp(\mathbf{z}_{aerial}^T \mathbf{z}_{bev}^{(j)} / \tau)}
\end{equation}

where $\tau$ is the temperature parameter (set to $0.07$) and $B$ is the batch size. The diagonal elements of the similarity matrix represent positive pairs, while off-diagonal elements serve as negatives. To improve training efficiency, we implement hard negative mining by selecting the top-$k$ hardest negatives (highest similarity) from the batch for each sample, focusing the learning on the most challenging negative pairs.

\begin{table*}[t]
    \begin{center}
    \resizebox{\textwidth}{!}{
        \setlength{\tabcolsep}{4pt}
        \begin{tabular}{ccccccccccccc}
            \toprule
            \multirow{2}[2]{*}{\textbf{Map}} &
            \multirow{2}[2]{*}{\textbf{Method}} &
            \multirow{2}[2]{*}{\begin{tabular}{c}\textbf{Map Size}  \\ (in pixels)\end{tabular}} &
            \multirow{2}[2]{*}{\begin{tabular}{c}\textbf{Avg. Loc. / Ori.}\\\textbf{Error} (m / $\degree$) $\downarrow$\end{tabular}} &
            \multicolumn{3}{c}{\textbf{Lat. R@Xm} $\uparrow$} &
            \multicolumn{3}{c}{\textbf{Long. R@Xm} $\uparrow$} &
            \multicolumn{3}{c}{\textbf{Ori. R@X$\degree$} $\uparrow$} \\
            \cmidrule(lr){5-7}\cmidrule(lr){8-10}\cmidrule(lr){11-13}
            & & & & 1m & 3m & 5m & 1m & 3m & 5m & 1$\degree$ & 3$\degree$ & 5$\degree$ \\
            \midrule
            \multirow{3}{*}{\begin{tabular}{c}OSM\end{tabular}} 
              & OrienterNet & $256\times 256$ & 28.01 / 8.30 & 2.86 & 8.80 & 15.37 & 4.03 & 11.47 & 18.48 & 8.62 & 24.11 & 36.38\\
              & AGL-Net      & $256\times 256$ & 18.02 / 8.59 & 6.55 & 18.88 & 31.17 & 5.16 & 15.00 & 24.54 & 6.25 & 17.94 & 31.87\\
            & \textbf{\ours} & $256\times 256$ & \first{15.32} / \first{9.21} & \first{8.35} & \first{21.42} & \first{34.76} & \first{7.32} & \first{19.23} & \first{29.12} & \first{9.53} & \first{25.13} & \first{39.45}\\
              
            \midrule
            \multirow{3}[1]{*}{\begin{tabular}{c}Satellite\end{tabular}}
              & OrienterNet* & $256\times 256$ & 27.75 / 8.65  & 3.44 & 9.53 & 16.89 & 7.48 & 14.21 & 19.62 & 9.26 & 26.23 & 38.21\\
              & AGL-Net*     & $256\times 256$ & 17.32 / 9.76 & 8.73 & 19.21 & 29.54 & 6.89 & 17.29 & 27.54 & 6.67 & 15.86 & 30.75\\
              & \textbf{\ours} & $256\times 256$ & \first{6.47} / \first{5.07} & \first{29.22} & \first{67.53} & \first{78.79} & \first{31.8} & \first{65.34} & \first{75.28} & \first{9.81} & \first{30.09} & \first{49.77}\\
              
            \bottomrule
        \end{tabular}
    }
    \end{center}
\caption{\small Results on the KITTI dataset using OSM and satellite aerial maps. The asterisk (*) indicates models that we retrained with satellite aerial maps to ensure consistency across baselines. TransLocNet significantly outperforms existing methods, reducing position error by over 60\% and achieving the most accurate orientation estimates in the challenging satellite setting.}
\label{tab:kitti_results}
\vspace{-3mm}
\end{table*}

\subsection{Multi-Objective Loss Function}

Our training objective combines multiple loss components to ensure robust localization:

\vspace{-10pt}
\begin{equation}
\mathcal{L}_{total} = \mathcal{L}_{loc} + \mathcal{L}_{ori} + \lambda_1 \mathcal{L}_{reg} + \lambda_2 \mathcal{L}_{contrastive}
\end{equation}

\textbf{Location Loss:} We employ a likelihood-based approach using KL divergence between predicted and target location distributions. This probabilistic formulation provides natural uncertainty quantification and handles ambiguous scenarios gracefully:

\vspace{-7pt}
\begin{equation}
\mathcal{L}_{loc} = \text{KL}(\mathcal{M}_{loc} || \mathcal{M}_{loc}^{gt})
\end{equation}

The KL divergence loss encourages the predicted distribution to match the ground truth while penalizing overconfident predictions in uncertain regions.

\textbf{Direct Regression Loss:} We apply Huber loss directly on predicted coordinates for additional supervision. The Huber loss provides robust training by being less sensitive to outliers compared to L2 loss, while maintaining smooth gradients compared to L1 loss:

\vspace{-5pt}
\begin{equation}
\mathcal{L}_{reg} = \text{Huber}((x^*, y^*), (x_{gt}, y_{gt}))
\end{equation}

\textbf{Orientation Loss:} The orientation loss uses KL divergence between predicted and target orientation distributions:

\vspace{-5pt}
\begin{equation}
\mathcal{L}_{ori} = \text{KL}(\mathcal{M}_{ori} || \mathcal{M}_{ori}^{gt})
\end{equation}

To determine optimal loss weights, we conduct systematic sensitivity analysis on the KITTI dataset. We evaluate different combinations of $\lambda_1 \in \{0.5, 1.0, 2.0, 4.0\}$ and $\lambda_2 \in \{0.05, 0.1, 0.2, 0.5\}$ while keeping the primary losses (location and orientation) at unit weight. Emperical results show that $\lambda_1 = 2.0$ and $\lambda_2 = 0.1$ provides the best performance. The higher weight for regression loss emphasizes direct coordinate supervision, while the lower weight for contrastive loss ensures it acts as a regularizer without dominating the primary localization objective.

\section{EXPERIMENTS AND RESULTS}

To validate the effectiveness of our proposed \ours framework, we conduct comprehensive experiments across multiple datasets and scenarios, demonstrating significant improvements in aerial-ground vehicle localization accuracy and robustness.

\subsection{Datasets}

We evaluate \ours on three diverse datasets that represent different scenarios and data modalities for aerial-ground vehicle localization:

\textbf{CARLA Dataset:} We utilize the synthetic CARLA dataset introduced in the AGL-Net paper \cite{guan2024agl}, which provides synchronized aerial and ground-level data collected by an ego vehicle in the CARLA simulation environment. The dataset encompasses multiple urban scenarios across different towns with varying driving conditions and scenarios. The aerial imagery has a scale of 1 pixel per meter.

\textbf{KITTI + OSM:} We employ the original KITTI dataset \cite{KITTI} combined with OpenStreetMap (OSM) data for map-based localization. The KITTI dataset provides real-world LiDAR point clouds collected from a vehicle equipped with a Velodyne laser scanner, along with precise GPS/INS ground truth poses. The OSM data serves as the reference map, containing vector-based road network information, building footprints, and other geographic features, with a scale of 2 pixels per meter.

\textbf{KITTI + Aerial Imagery:} We extend the KITTI dataset by incorporating aerial imagery as the map reference, following the approach described in \cite{HighlyAccurate-shi}. This dataset combines the original KITTI LiDAR point clouds with high-resolution aerial imagery at a scale of 5 pixels per meter. Unlike CARLA, the aerial imagery is not temporally aligned with LiDAR collection, introducing temporal and spatial misalignment challenges. This represents the most challenging scenario due to viewpoint differences, modality gaps, and lack of temporal synchronization.

\subsection{Metrics}

We evaluate our method using both recall and metric distance for evaluation.

\textbf{Recall@Xm/\degree:} This metric measures the percentage of predicted poses that fall within $X$ meters or $X$ degrees of the ground-truth poses. Specifically, position recall is reported at thresholds of $1$m, $3$m, and $5$m, while orientation recall is evaluated at $1$\degree, $3$\degree, and $5$\degree.  

\textbf{Metric Distance:} This metric reports the average absolute error between the predicted and ground-truth poses, expressed separately in meters for position and degrees for orientation.  

\subsection{Implementation Details}

\ours is implemented in PyTorch. Both the aerial and BEV encoders are based on ResNet-34 backbones with multi-scale feature fusion and lightweight spatial/channel attention. The encoders output features with dimension $128$, which are then processed by a cross-modal attention module with $4$ heads to align aerial and BEV features. The likelihood decoder outputs both position and orientation likelihood maps, with orientation discretized into 360 bins.  

For contrastive learning, we apply two-layer projection heads that map the 128-dimensional features to 64-dimensional embeddings, followed by an InfoNCE loss with temperature $\tau=0.07$. Models are trained using the Adam optimizer with an initial learning rate of $1\times10^{-4}$. All experiments are conducted on an NVIDIA A100 ($80$GB) GPU.

\textbf{Computational complexity:} The computational complexity of \ours is primarily determined by the cross-modal attention mechanism and feature extraction. The attention computation has complexity $O(N^2 \cdot d)$ where $N$ is the number of spatial locations and $d$ is the feature dimension. With our configuration ($N = 256 \times 256$ and $d = 128$), the model processes aerial images of size $256 \times 256$ and BEV maps of the same resolution. The total model size is approximately $45$M parameters.

\textbf{Training strategy:} Given the limited size of the KITTI dataset compared to the synthetic CARLA dataset, we employ comprehensive data augmentation strategies specifically for KITTI training. Our augmentation pipeline includes random aerial map rotation (up to $\pm60\degree$ with 70\% probability), random map cropping around the vehicle position, and color jittering. These augmentations help the model generalize to various weather and lighting conditions.

\subsection{Results}

\textbf{Qualitative results:} Fig. \ref{fig:vis_carla} and \ref{fig:vis_kitti} illustrate qualitative localization performance on the CARLA and KITTI datasets, respectively. On CARLA dataset, where aerial and ground-level data are synchronized and noise-free, \ours achieves precise pose estimation, typically with sub-meter translation error and sub-degree orientation error. On KITTI dataset, real-world conditions introduce noise and temporal misalignment between aerial imagery and LiDAR scans, yet the model generates concentrated likelihood heatmaps and reasonable predictions, keeping errors within practical bounds. These results demonstrate both the accuracy of our method in ideal conditions and its robustness to real-world challenges.

\textbf{Quantitative results.} 
Table~\ref{tab:carla_results} summarizes performance on the CARLA dataset, comparing with two baselines: AGL-Net \cite{guan2024agl} and OrienterNet \cite{sarlin2023orienternet}. These baselines are the most recent and relevant approaches for aerial-ground (LiDAR) localization, ensuring fair comparison under identical experimental conditions. At $256\times256$ resolution ($\sim$256 meters), \ours achieves an average error of 1.57 m / 3.22$\degree$, improving over AGL-Net (1.83 m / 3.76$\degree$) and outperforming OrienterNet (2.23 m / 19.15$\degree$) in orientation prediction. In addition, \ours attains \textbf{100\% recall} at 5 m, indicating reliable predictions across different weather and lighting conditions. At the higher $1024\times1024$ resolution ($\sim$1 kilometer), overall errors naturally increase due to the larger search space, yet \ours still achieves notable gains. Compared to AGL-Net, \ours reduces the localization error by \textbf{62\%} (5.75 m vs. 14.96 m) and the orientation error by \textbf{80\%} (10.93$\degree$ vs. 57.25$\degree$).

Table~\ref{tab:kitti_results} presents results on KITTI with map size of $256 \times 256$ for both OSM  ($\sim$512 meters) and aerial maps ($\sim$1280 meters). For OSM-based localization, \ours achieves the lowest position error of 15.32 m and a comparable orientation error of 9.21$\degree$ relative to the baselines. In the more challenging aerial setting, the improvements are more pronounced: \ours obtains the lowest errors for both position and orientation. Specifically, \ours reduces the position error to 6.47 m, a 63\% improvement over the runner-up AGL-Net. Orientation accuracy also improves to 5.07$\degree$, compared to 9.76$\degree$ (AGL-Net) and 8.65$\degree$ (OrienterNet).

In terms of location and orientation recall at different thresholds, \ours consistently outperforms the baselines by large margins. These gains demonstrate the effectiveness of our model design in bridging modality and viewpoint gaps, especially in the aerial setting where image-only methods tend to struggle.

\subsection{Ablation Studies}

We conduct ablation studies to validate our design choices by evaluating three key components: (1) cross-modal attention versus simple concatenation, (2) multiscale versus single-scale BEV feature extraction, and (3) probabilistic decoder versus direct regression. Table~\ref{tab:abl_components} presents results on KITTI with $256 \times 256$ map size.

\begin{table}[t]
    \begin{center}
      \resizebox{\linewidth}{!}{
          \setlength{\tabcolsep}{1pt} 
          \begin{tabular}{cccccccc}
              \toprule
              \multirow{2}[2]{*}{
              \begin{tabular}{c}
                  \textbf{Cross} \\ \textbf{attn}
              \end{tabular}
              } &   
              \multirow{2}[2]{*}{
              \begin{tabular}{c}
                  \textbf{MultiScale.} \\ \textbf{features}
              \end{tabular}
              } &
              \multirow{2}[2]{*}{
              \begin{tabular}{c}
                  \textbf{Prob.} \\ \textbf{decoder}
              \end{tabular}
              } &
              \multirow{2}[2]{*}{
              \begin{tabular}{c}
                  \textbf{Avg. Loc. / Ori.}\\ \textbf{Error} (m / \degree) $\downarrow$
              \end{tabular}
              } & \multicolumn{2}{c}{\textbf{Loc. R@Xm} $\uparrow$} &  \multicolumn{2}{c}{{ \textbf{Ori. R@X\degree $\uparrow$}
              }}  \\ \cmidrule(lr){5-6} \cmidrule(lr){7-8}              
                  &  &   &  & 2m & 5m & 2\degree & 5\degree \\
               \midrule
               
              \redx & \greencheck & \greencheck & 15.37 / 11.25 & 22.35 & 47.96 & 16.24 & 38.56\\
              
              \greencheck & \redx & \greencheck & 7.84 / 7.92 & 26.85 & 59.12  & 16.68 & 43.29 \\
              \greencheck & \greencheck & \redx & 8.73 / 7.16 & 27.34 & 57.78 
           & 18.14 & 45.26\\
              \greencheck & \greencheck & \greencheck & \first{6.47} / \first{5.07} & \first{32.60} & \first{67.13}  & \first{20.81} & \first{49.77}\\
          
            \bottomrule
          \end{tabular}
      }
      \end{center}
  \caption{\small Ablation studies on model components. For cross-modal attention, we replace it with simple concatenation of aerial and BEV features. For the multiscale feature extraction, we compare models with multiscale BEV feature extraction enabled versus disabled. For the likelihood decoder, we replace it with direct $(x,y,\theta)$ regression to predict pose coordinates directly. Results are reported on the KITTI dataset.}
  \label{tab:abl_components}
  \vspace{-3mm}
  \end{table}

The ablation results demonstrate the contribution of each component. Removing cross-modal attention causes the most severe degradation with errors of 15.37 m / 11.25$\degree$ and low recall rates (22.35\% @ 2m, 16.24\% @ 2\degree), indicating that simple concatenation fails to capture complex spatial relationships between aerial and BEV features.

Removing multiscale BEV feature extraction leads to drops in the model's performance. Replacing the probabilistic decoder with direct regression also shows similar moderate degradation with 8.73 m / 7.16$\degree$ errors. With all three components enabled, \ours achieves the best results across all metrics.

\begin{table}[t]
    \begin{center}
      \resizebox{\linewidth}{!}{
          \setlength{\tabcolsep}{1pt} 
          \begin{tabular}{cccccccc}
              \toprule
              \multirow{2}[2]{*}{
              \begin{tabular}{c}
                  \textbf{Direct reg.} \\ \textbf{loss}
              \end{tabular}
              } &   
              \multirow{2}[2]{*}{
              \begin{tabular}{c}
                  \textbf{Contrastive} \\ \textbf{loss}
              \end{tabular}
              } &
              \multirow{2}[2]{*}{
              \begin{tabular}{c}
                  \textbf{Use Fourier} \\ \textbf{features}
              \end{tabular}
              } &
              \multirow{2}[2]{*}{
              \begin{tabular}{c}
                  \textbf{Avg. Loc. / Ori.}\\ \textbf{Error} (m / \degree) $\downarrow$
              \end{tabular}
              } & \multicolumn{2}{c}{\textbf{Loc. R@Xm} $\uparrow$} &  \multicolumn{2}{c}{{ \textbf{Ori. R@X\degree $\uparrow$}
              }}  \\ \cmidrule(lr){5-6} \cmidrule(lr){7-8}              
                  &  &   &   & 2m & 5m & 2\degree & 5\degree \\
               \midrule
               
              \redx & \greencheck & \greencheck & 7.15 / 6.54 & \first{33.11} & 66.35 & 18.63 & 45.11\\
              
              \greencheck & \redx & \greencheck & 8.54 / 7.47 & 29.24 & 59.76  & 17.25 & 42.59 \\
              \greencheck & \greencheck & \redx & 6.82 / 7.34 & 32.71 & 65.57  & 16.21 & 41.75 \\
              \greencheck & \greencheck & \greencheck & \first{6.47} / \first{5.07} & 32.60 & \first{67.13}  & \first{20.81} & \first{49.77} \\
          
            \bottomrule
          \end{tabular}
      }
      \end{center}
  \caption{\small Ablation of the loss function components and Fourier features on KITTI. We evaluate the contributions of direct regression loss, contrastive loss, and Fourier features to the total performance.}
  \label{tab:abl_loss}
  \vspace{-3mm}
  \end{table}

We further evaluate the impact of the direct regression loss, contrastive loss, and Fourier features in Table~\ref{tab:abl_loss}. For the Fourier features ablation, we compare models with and without the FFT layer applied to BEV features, where the FFT layer computes the log-amplitude spectrum to provide rotation-invariant representations. The direct regression loss alone yields moderate improvements while accelerating convergence. The contrastive loss acts as a regularizer, lowering error and enhancing recall. Fourier features particularly help with orientation prediction accuracy by encoding periodic patterns in the aerial imagery. The combination of all three components achieves the best overall performance, producing the lowest errors and highest recall across all metrics.

\subsection{Limitations and Future Work}

While \ours improves aerial-ground vehicle localization, limitations include dependence on high-quality aerial imagery, sensitivity to LiDAR preprocessing, and computational complexity that may hinder real-time deployment. Future work will focus on developing robust features for low-quality data, lightweight attention methods, dynamic scene handling, and incorporating additional sensors like IMUs and aerial LiDAR.

\section{CONCLUSION}

In this work, we present \ours, a cross-modal attention framework for aerial–ground vehicle localization that bridges the viewpoint and modality gaps between ground-level LiDAR point clouds and overhead aerial imagery. Our probabilistic design employs cross-modal attention to dynamically align BEV projections of LiDAR data with aerial features, complemented by a contrastive learning module that improves feature discriminability across modalities. \ours consistently outperforms existing baselines, as shown by extensive experiments on both synthetic and real-world datasets. These results demonstrate the robustness and generalization of \ours for reliable aerial–ground localization in both controlled and challenging real-world environments, marking a step forward for autonomous navigation systems.

\section*{ACKNOWLEDGMENTS}
This material is based upon work supported in part by the DEVCOM Army Research Laboratory under cooperative agreement : W911NF2520170.









{\small
\bibliographystyle{IEEEtran}
\bibliography{refs}
}

\end{document}